%% file: main.tex
\documentclass[journal, 10pt]{IEEEtran} 

\IEEEoverridecommandlockouts                                
\usepackage{mathtools,amssymb,commath}
\usepackage{amsmath}
\usepackage{graphicx,import}
\usepackage{verbatim}
\usepackage{color}
\usepackage{hyperref}
\hypersetup{colorlinks=false}  
\usepackage{subcaption}
\usepackage{microtype}

\DeclareMathOperator*{\minimize}{minimize}

\usepackage{mathtools,amsthm}

\begin{document}

\title{Jerk Control of Floating Base Systems \\ with Contact-Stable Parametrised Force Feedback}

\author{Ahmad Gazar$^{\textbf{*},1}$, Gabriele Nava$^{\textbf{*},2,3}$, Francisco Javier Andrade Chavez$^{2}$, Daniele Pucci$^{2}$ 

\thanks{$^{\textbf{*}}$ \textbf{The two authors equally contributed to this paper.}}%
\thanks{Manuscript received July 22, 2019; revised February 15, 2020; accepted April 23, 2020.}%
\thanks{This paper is supported by EU An.Dy Project. This project has received funding from the European Union’s Horizon $2020$ research and innovation programme under grant agreement No. $731540$.  The content of this publication is the sole responsibility of the authors.  The European Commission or its services cannot be held responsible for any use that may be made of the information it contains.}%
\thanks{$^{1}$ Max Planck Institute for Intelligent Systems, T\"ubingen, Germany, {\tt\small name.surname@tuebingen.mpg.de}}%
\thanks{$^{2}$ Dynamic Interaction Control, Istituto Italiano di Tecnologia, Genova, Italy {\tt\small name.surname@iit.it}}%
\thanks{$^{3}$ DIBRIS, University of Genova, Genova, Italy}%
\thanks{Digital Object Identifier (DOI): see top of this page.}
\thanks{© © 2020 IEEE. Personal use of this material is permitted. Permission from IEEE must be obtained for all other uses, in any current or future media, including reprinting/republishing this material for advertising or promotional purposes, creating new collective works, for resale or redistribution to servers or lists, or reuse of any copyrighted component of this work in other works.}
}

\markboth{IEEE Transactions on Robotics, Preprint Version. Accepted September, 2020}%
{Shell \MakeLowercase{\textit{Gazar, Nava et al.}}: Jerk Control of Floating Base Systems with Contact-Stable Parametrised Force Feedback}

\maketitle

\begin{abstract}
Nonlinear controllers for floating base systems in contact with the environment are often framed as quadratic programming (QP) optimization problems. Common drawbacks of such QP based controllers are: 
the control input often experiences discontinuities; no force feedback from Force/Torque (FT) sensors installed on the robot is taken into account. This paper attempts to address these limitations 
using \emph{jerk} based control architectures. 
The proposed controllers assume the rate-of-change of the joint torques as control input, and exploit the system position, velocity, accelerations, and contact wrenches as measurable quantities. The key ingredient of the presented approach is a one-to-one correspondence between free variables and 
an 
inner approximation of 
the manifold defined by the contact stability constraints. 
More precisely, the proposed correspondence covers completely the contact stability manifold except for the so-called friction cone, for which there exists a unique correspondence for more than 90\% of its elements.  
The correspondence  allows us to transform the underlying constrained optimisation problem into one that is unconstrained. Then, we propose a \emph{jerk} control framework that exploits the proposed correspondence and uses FT measurements in the control loop. Furthermore, we present Lyapunov stable controllers for the system momentum in the \emph{jerk} control framework. The  approach is validated with simulations and experiments using the iCub humanoid robot.

\end{abstract}


%
\IEEEpeerreviewmaketitle

\import{tex/}{intro}
\import{tex/}{background}
\import{tex/}{statement}
\import{tex/}{parametrization}
\import{tex/}{control}
\import{tex/}{results}
\import{tex/}{conclusions}
\import{tex/}{appendix}

\addtolength{\textheight}{0cm}     

\ifCLASSOPTIONcaptionsoff
  \newpage
\fi

%

\bibliographystyle{IEEEtran}
\bibliography{IEEEabrv,Biblio}

%

\begin{IEEEbiography}[{\includegraphics[width=1in,trim={0 0 0 0},clip,keepaspectratio]{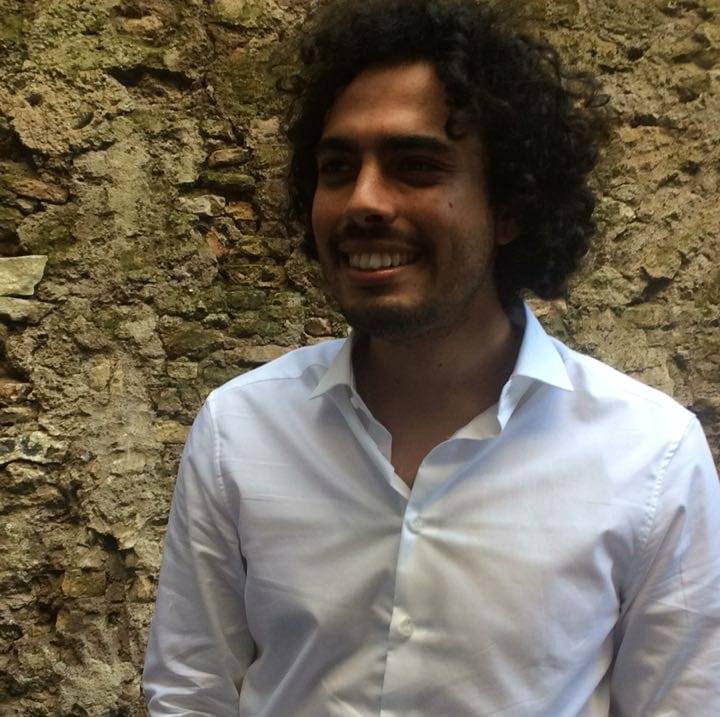}}]{Ahmad Gazar}
Ahmad received his Bachelor degree in 2011 in Engineering and Material Science from the German University in Cairo. He received his master degree in 2018 in Artificial Intelligence and Robotics from La Sapienza University in Rome, after he conducted his master thesis in Dynamic Interaction and Control lab at the Italian Institute of Technology (IIT) in Genova under the supervision of Daniele Pucci. Currently, he is pursuing his PhD in Robust Model Predictive Control for Locomotion of Legged Robots, at the Movement Generation and Control lab, at Max-Planck Institute for Intelligent Systems in Tuebingen under the supervision of Ludovic Righetti and Andrea Del Prete. His main research interests are optimal control, and trajectory optimization for robots with arms and legs using robust and stochastic MPC (Model Predictive Control).
\end{IEEEbiography}

\vspace{-0.5cm}

\begin{IEEEbiography}[{\includegraphics[width=1in,trim={170 0 160 0},clip,keepaspectratio]{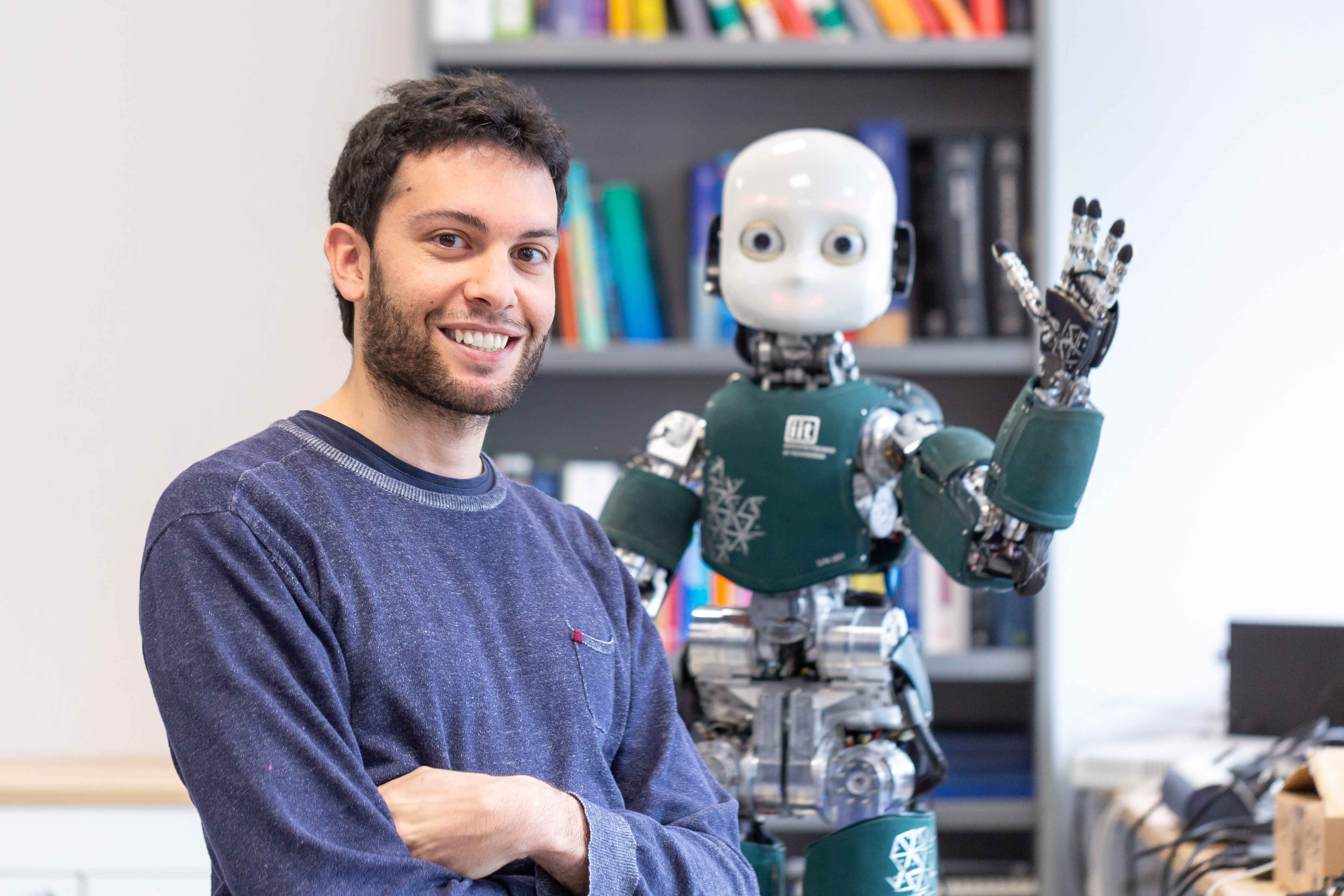}}]{Gabriele Nava}
Gabriele received the Bachelor and Master degrees in Mechanical Engineering from Politecnico di Milano, in 2013 and 2015, respectively. In 2020 he earned the PhD title in Bioengineering and Robotics at Universit\`{a} degli Studi di Genova, in cooperation with the Italian Institute of Technology (IIT), under the supervision of Daniele Pucci and Giorgio Metta. He is currently working as postdoc researcher in the Dynamic Interaction Control Lab at IIT. His main research topic concerns the design of control algorithms for floating base systems, with a focus on Aerial Humanoid Robotics. He is the Scrum Master of the iRonCub group, which pursues the objective of making the humanoid robot iCub fly. 
\end{IEEEbiography}

\vspace{-0.5cm}

\begin{IEEEbiography}[{\includegraphics[width=1in,trim={170 0 160 0},clip,keepaspectratio]{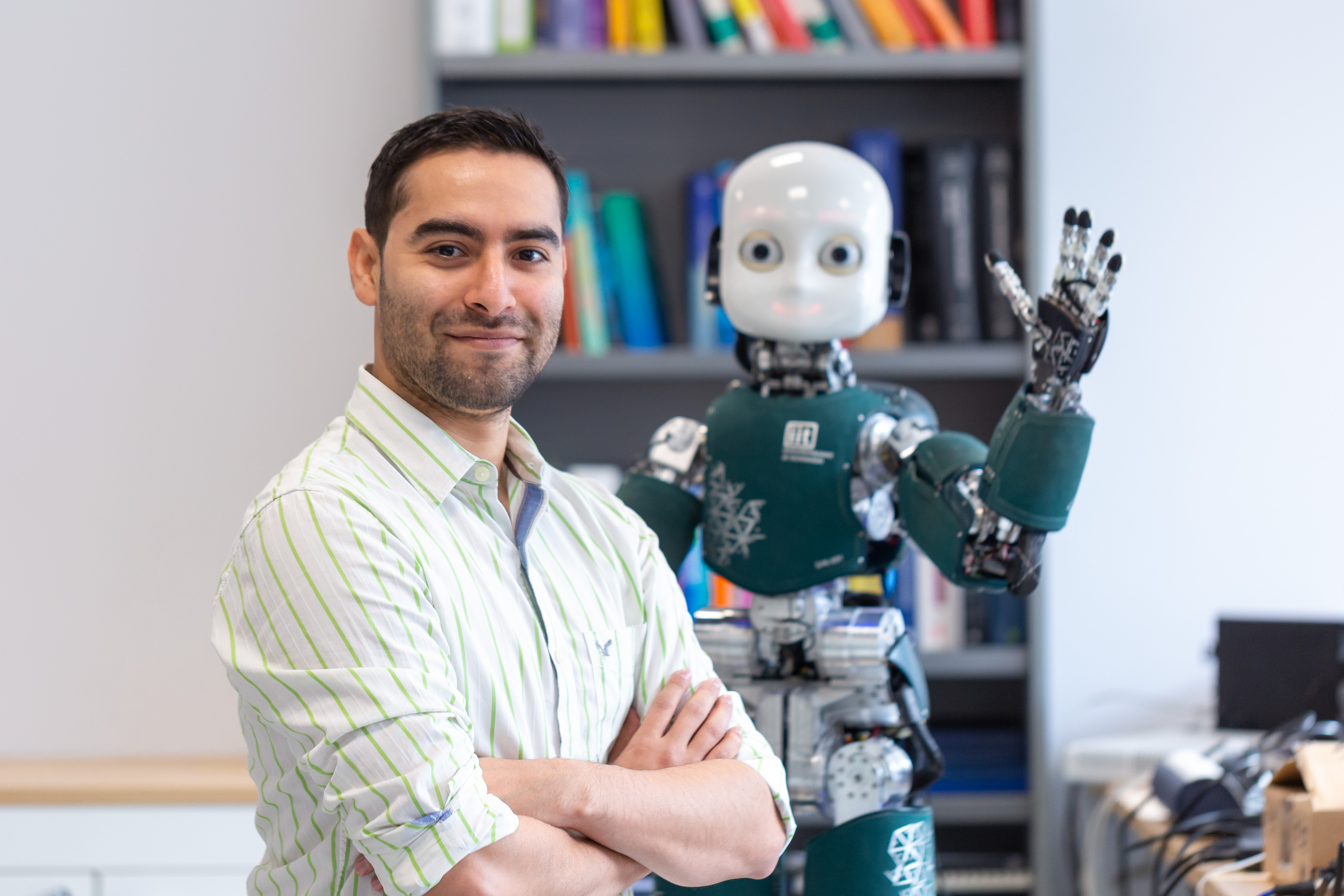}}]{Francisco Andrade}
Francisco received his Bachelor degree in Mechantronics Engineering from Universidad Anahuac Mayab, Merida, Mexico in 2011. He received a double Master Degree in Advance Robotics from Universit\`{a} degli Studi di Genova and Ecole Central de Nantes in 2015. He successfully concluded a PhD in Bioengineering and Robotics at Universit\`{a} degli Studi di Genova in cooperation with the Istituto Italiano di Tecnologia (IIT) under the supervision of Daniele Pucci in 2019. From 2019 to 2020, he was a Postdoc in the Dynamic Interaction Control Lab at IIT. During this period, he also held a Scrum Master position, first for the Telexistence group and later for an Industrial Collaboration Project. Francisco Javier Andrade Chavez is currently a PostDoc and lab manager in the Human-Centered Robotics Lab at University of Waterloo, Canada. His main research interests are sensing, estimation, and control of dynamics and its application to bio-inspired robots and human-robot collaboration.
\end{IEEEbiography}

\vspace{-0.5cm}

\begin{IEEEbiography}[{\includegraphics[width=1in,height=1.25in,clip,keepaspectratio]{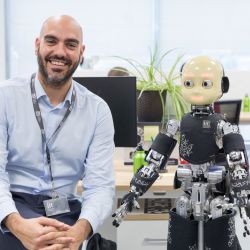}}]{Daniele Pucci}
Daniele received the bachelor and master degrees in Control Engineering with highest honors from ”Sapienza”, University of Rome, in 2007 and 2009, respectively. In 2009, he also received the ”Academic Excellence Award” from Sapienza. In 2013, he earned the PhD title in Nonlinear control applied to Flight Dynamics, with a thesis prepared at INRIA Sophia Antipolis, France, under the supervision of Tarek Hamel and Claude Samson. From 2013 to 2017, he has been a postdoc at the Istituto Italiano di Tecnologia (IIT) working within the EU project CoDyCo. Since August 2017, he is the head of the Dynamic Interaction Control lab.  The main lab research focus is on the humanoid robot locomotion problem, with specific attention on the control and planning of the associated nonlinear systems. Also, the lab is pioneering Aerial Humanoid Robotics, whose main aim is to make flying humanoid robots. Currently, the lab is implementing iRonCub, the jet-powered flying version of the humaonid robot iCub. Daniele is also the scientific PI of the H2020 European Project AnDy, task leader of the H2020 European Project SoftManBot, and coordinator of the joint laboratory between IIT and Honda JP. In 2019, he was awarded as Innovator of the year Under 35 Europe from the MIT Technology Review magazine. Since 2020 and in the context of the split site PhD supervision program, Daniele is a visiting lecturer at University of Manchester.
\end{IEEEbiography}




\end{document}

%% file: tex/intro.tex
\section{Introduction}
\label{sec:intro}

\IEEEPARstart{N}{onlinear} feedback control of fixed-based (e.g. manipulators) and floating-base (e.g. humanoids) robots is not new to the Control community~\cite{deWit1996,yoshikawa2000,mistry2010,sentis2005}. Feedback-linearisation, robust control, and adaptive laws are only few examples of the large variety of control methods developed for steering these  systems towards desired quantities. When fixed and floating-base systems make contact with the environment, the robot control  has to deal with  regulating also the  forces and torques towards values that ensure a desired interaction. This paper contributes towards the stabilisation of floating-base systems\renewcommand{\thefootnote}{\Roman{footnote}}
in contact with the environment by proposing controllers that ensure \emph{contact stability} (see, e.g., \cite{Frontiers2015}) while including 
force
feedback  
in the control laws. The proposed  approach 
uses
the rate-of-change of the joint torques 
as  control input, 
and for this reason 
it is here referred to as \emph{jerk} control. 


Force control strategies for fixed-base systems can be roughly divided into two categories: \emph{direct} and \emph{indirect} force control methods~\cite{deWit1996,Villani2015}. Indirect methods achieve compliance without explicitly closing a feedback loop on the measured contact forces. 
%
In particular, \emph{impedance control} is a common objective for indirect techniques whose goal is often that of achieving a desired dynamic behavior of the robot's end-effector. The control design requires to include force and torque measurements as feedforward terms to achieve full feedback linearization of the end-effector dynamics. If no force measurement is available, a simplified \emph{stiffness control} can still be applied \cite{deWit1996}.
%
On the other hand, direct force control methods include explicit feedback from the measured interaction wrenches, usually related to the force error \cite{Villani2015,ortenzi2017}. An example of these techniques is the hybrid position/force control, which is often applied when the environment is rigid and the end-effector has to continuously maintain a contact. The rigid environment assumption enables the decomposition of the end-effector dynamics into a \emph{constrained} and a \emph{free} direction~\cite{raibert1981}. Along the constrained directions, feasible desired forces are exerted. Here, additional feedback terms are added to ensure convergence in presence of external disturbances and unmodeled dynamics. Although the \emph{desired} force satisfies the contact stability constraints, the \emph{commanded} force, which includes the feedback from Force/Torque (FT) sensors, may instantaneously violate the contact constraints.

The recent research effort on humanoid robots gave impetus to the force control of floating-base systems~\cite{Featherstone2007,Ott2011,Wensing2013,Hopkins2015a}.  These systems are often underactuated, namely, the number of control inputs is fewer than the system’s degrees of freedom~\cite{spong1998}. 
The lack of actuation is usually circumvented by means of the contacts between the robot and the environment, but this requires close attention to the
forces the robot exerts at the contact locations. If not regulated appropriately, uncontrolled contact forces may break a contact, which makes the robot control critical~\cite{Ott2011,Wensing2013}.
%
The contacts between a floating-base system and the environment are often assumed to be rigid and \emph{planar}, although 
compliant contacts and uneven ground are also considered in the literature \cite{Azad2015,Caron2019,Henze2018,liu2015b}.
Furthermore, all contacts are also \emph{unilateral}, being the robot not physically attached to  ground and in general able to make and break contacts. Contact activation and deactivation occur continuously, e.g. in the case of humanoid robot walking, and can be addressed with the design of a proper \emph{state machine} that plans references for a balancing/walking controller \cite{park2013,liu2015}. 
From the control design perspective, a common strategy for floating-base systems is based on the so called stack-of-task approach~\cite{mansard2009}. These strategies usually consider several control objectives organized in a hierarchical or weighted prioritization. Often, the high-priority task is the stabilization of the robot momentum~\cite{koolen2015design}: 
its objective is 
the control of the robot center-of-mass and angular momentum by means of contact forces while guaranteeing stable zero-dynamics~\cite{nava2016}. Quadratic programming (QP) solvers can be used to monitor contact forces to ensure both robot and contact stability \cite{Lee2012}. 


Analogously to 
direct and indirect force controllers, QP based force control of floating base systems usually suffers from the following limitations:
%
$i)$  optimal control inputs (i.e., the joint torques) may be discontinuous in certain cases, e.g. during the switching from two feet and one foot balancing of humanoid robots. In this case, the robot control may become critical.
Although  further (and often numerous)
constraints can be added to the QP solver to enforce continuity, the effectiveness of this approach is not always satisfactory in practice; 
$ii)$
force feedback from FT sensors is missing in the control action. 

This paper proposes a control approach that attempts to address these 
two
main limitations of QP based controllers for floating base systems. The key ingredients of our approach are: $a)$ propose an invertible one-to-one mapping between a set of constraint-free variables and 
an 
inner approximation of the 
contact stability manifold. 
$b)$  propose controllers that use the wrench mapping and exploit the rate-of-change of the joint torques as control input; $c)$ extend the proposed controllers to the case when 
the joint torques (and not their rate-of-change) is the available control input. The proposed approach exploits a relative degree augmentation of the underlying system, i.e., the system state is composed of the system position, velocity, and acceleration. For this reason, the proposed approach is referred to as \emph{jerk} control, which also incorporates force feedback from FT sensors. 
Furthermore, 
we 
present control laws that stabilise a desired robot momentum using joint-torques as input and having guaranteed Lyapunov stability properties. 
%
The proposed approach is validated with simulations and experiments using the humanoid robot iCub balancing on rigid contacts.

The paper is organized as follows. Section \ref{sec:background} recalls the notation, the robot modeling, the contact stability constraints, and introduces the problem statement. Section~\ref{sec:parametrization} presents a novel contact-stable parametrization of the contact wrenches. Section~\ref{sec:control} introduces the main ideas behind \emph{jerk} control using the contact-stable parametrisation. Section~\ref{sec-jerk-momentum-control} presents Lyapunov stable \emph{jerk} controllers when the control objective is the stabilisation of the robot momentum. Section~\ref{sec:results} presents validations of the approach on the humanoid robot iCub. Conclusions and perspectives conclude the paper. 

%% file: tex/background.tex
\section{Background}
\label{sec:background}

\subsection{Notation}

\begin{itemize}
  \item $\mathcal{I}$ is an inertial frame, with its $z$ axis pointing against the gravity, $\mathcal{B}$ is a frame attached to the robot's \emph{base link}.
  \item The constant $m$ represents the total mass of the robot, and $g$ is the norm of the gravitational acceleration.
  \item Given a matrix $A \in \mathbb{R}^{p \times n}$, we denote with $A^{\dagger} \in \mathbb{R}^{n \times p}$ its Moore-Penrose pseudoinverse.
  \item $e_i \in \mathbb{R}^6$ is the canonical vector, consisting of all zeros but the $i$-th component that is equal to one.
  \item Let $S(x) \in \mathbb{R}^{3 \times 3}$ be the skew-symmetric matrix such that $S(x)y {=} x {\times} y$, with $\times$ the cross product 
  operator 
  in $\mathbb{R}^3$. 
\end{itemize}

\subsection{Robot Modeling}
The robot is modeled as a multi-body system composed of $n + 1$ rigid bodies, called links, connected by $n$ joints with one degree of freedom each. We also assume that none of the links has an \emph{a priori} constant position-and-orientation with respect to an inertial frame, i.e. the system is \emph{free floating}. 


The robot configuration space is the Lie group $\mathbb{Q} = \mathbb{R}^3 \times \mathbb{SO}(3) \times \mathbb{R}^n$ and it is characterized by the \emph{pose} (i.e. position-and-orientation) of a \emph{base frame} attached to the robot's \emph{base link}, and the joint positions. An element $q \in \mathbb{Q}$ can be defined as the following triplet: $q = (^{\mathcal{I}}o_{\mathcal{B}}, {}^{\mathcal{I}}R_{\mathcal{B}}, s)$ where $^{\mathcal{I}}o_{\mathcal{B}} \in \mathbb{R}^3$ denotes the origin of the base frame with respect to the inertial frame, $^{\mathcal{I}}R_{\mathcal{B}}$ is the rotation matrix representing the orientation of the base frame, and $s \in \mathbb{R}^n$ is the joint configuration characterizing the robot posture. The velocity of the multi-body system can be characterized by the \emph{algebra} of $\mathbb{Q}$. We here choose to represent the velocity of the multi-body system by the set $\mathbb{V} = \mathbb{R}^3 \times \mathbb{R}^3 \times \mathbb{R}^n$, where an element of $\mathbb{V}$ is the vector $\nu = (^{\mathcal{I}}\dot{o}_{\mathcal{B}}, \,^{\mathcal{I}}\omega_{\mathcal{B}},\, \dot{s}) =(\mathrm{v}_{\mathcal{B}}, \dot{s})$, and $^{\mathcal{I}}\omega_{\mathcal{B}}$ is the angular
velocity of the base frame expressed w.r.t. the inertial frame, i.e. $^{\mathcal{I}}\dot{R}_{\mathcal{B}} = S(^{\mathcal{I}}\omega_{\mathcal{B}})^{\mathcal{I}}R_{\mathcal{B}}$. A more detailed description of the configuration space and its algebra is provided in \cite{traversaro2017}. 

We assume that the robot interacts with the environment by exchanging $n_c$ distinct wrenches. The equations of motion of the multi-body system can be described by applying the Euler-Lagrange formalism \cite[Chapter 13.5]{Marsden1999}:
\begin{IEEEeqnarray}{LCL}
\IEEEyesnumber
  \label{eq:dynamics}
  M(q)\dot{\nu} + C(q,\nu)\nu + G(q) = B\tau + \sum_{k=1}^{n_c} J_{C_k}^T f^k,
\end{IEEEeqnarray}
where $M \in \mathbb{R}^{n+6 \times n+6}$ is the mass matrix, $C \in \mathbb{R}^{n+6 \times n+6}$ is the matrix accounting for Coriolis and centrifugal effects, $G$ $\in \mathbb{R}^{n+6}$ is the gravity term, $B = [0_{n\times6},\,\, 1_n]^\top$    is a selector of the actuated DoFs, $\tau \in \mathbb{R}^n$ is a vector representing the internal actuation torques, and $f^k \in \mathbb{R}^6$ represents an external wrench applied by the environment to the link of the $k$-th contact. The Jacobian $J_{\mathcal{C}_k} = J_{\mathcal{C}_k}(q)$ is the map between the robot's velocity $\nu$ and the linear and angular velocity of the $k$-th contact link. 

Lastly, it is assumed that a set of holonomic constraints acts on Eq. \eqref{eq:dynamics}
of the form $c(q) = 0$: they may represent, for instance, a frame having a constant position-and-orientation w.r.t. the inertial frame. 
Hence, we represent the holonomic constraints on links in rigid contact with the environment as $J_{\mathcal{C}_k}(q)\nu = 0$. The holonomic constraints associated with all the rigid contacts can be then compactly represented as:
\begin{IEEEeqnarray}{LCL}
\IEEEyesnumber
  \label{eq:holonomic_constraint}
  J(q) \nu = & \begin{bmatrix}
                 J_{\mathcal{C}_1}(q)\\
                 ...\\
                 J_{\mathcal{C}_{n_c}}(q)
               \end{bmatrix} \nu = 0.
\end{IEEEeqnarray}

By differentiating the kinematic constraints Eq. \eqref{eq:holonomic_constraint}, one has:
\begin{IEEEeqnarray}{LCL}
\IEEEyesnumber
  \label{eq:holonomic_constraint_acc}
  J\dot{\nu} + \dot{J}\nu = 0.
\end{IEEEeqnarray}
By combining the system dynamics \eqref{eq:dynamics} and the constraint equations 
\eqref{eq:holonomic_constraint_acc}, one obtains 
the following set of equations:
\begin{IEEEeqnarray}{LCL}
 \IEEEyesnumber \label{eq:constrained_dynamics}
  M\dot{\nu} + h &=&\, J^\top f + B\tau  \IEEEyessubnumber \label{eq:dynamics_constr} \\ 
  J\dot{\nu} + \dot{J}\nu &=& 0, \IEEEyessubnumber \label{eq:constraint_acc}
\end{IEEEeqnarray}
where 
$h:= C(q,\nu)\nu + G(q)$, and $f:= [f^1;...; f^{n_c}] \in \mathbb{R}^{6n_c}$ is the 
vector
of contact wrenches 
making Eq. \eqref{eq:holonomic_constraint_acc} satisfied.


\subsection{Contact Stability Constraints}
\label{sub-section:contact-stability-constraints}

\begin{figure}[!t]
\includegraphics[height=5cm,width=\columnwidth]{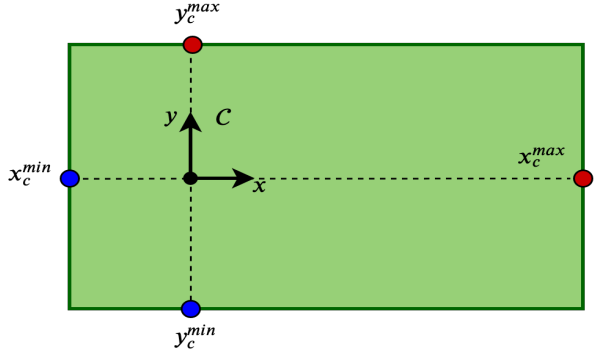}
  \centering 
  \caption{Contact surface. The picture highlights the rectangle's dimensions w.r.t. the contact frame $\mathcal{C}$.}
  \label{fig:foot_size}
\end{figure}

Often, the holonomic constraints acting on the system represent a robot flat surface in \emph{complete} rigid contact with the environment. To maintain the constraints \emph{stable}, then, conditions on the contact forces and moments shall be met. More precisely, let $f^k \in \mathbb{R}^6$ denote the vector of the $k$-th contact wrench associated with the $k$-th  \emph{active} rigid contact of a planar robot surface, namely $f^k= [f_x, \hspace{1 mm} f_y, \hspace{1 mm} f_z, \hspace{1 mm} M_x, \hspace{1 mm} M_y, \hspace{1 mm} M_z]^\top$. Then, the  \emph{contact stability} conditions to preserve
the planar unilateral contact
can be formulated as follows:
\begin{IEEEeqnarray}{LCL}
\IEEEyesnumber
  \label{eq:contact_constraints}
  f_z \hspace{1 mm} > \hspace{1 mm}  f_z^{min} \geq 0    \IEEEyessubnumber \label{eq:pos_vert_forces},\\ 
  \sqrt[]{f_x^2 + f_y^2} \hspace{1 mm}  < \hspace{1 mm} \mu_c f_z \IEEEyessubnumber \label{eq:static_friction},\\ 
 y_c^{min} < \frac{M_x}{f_z} < y_c^{max} \IEEEyessubnumber \label{eq:CoP_y},\\ 
 x_c^{min} < -\frac{M_y}{f_z} < x_c^{max} \IEEEyessubnumber \label{eq:CoP_x},\\ 
  \left |\frac{M_z}{f_z} \right|  \hspace{1 mm}  < \hspace{1 mm}  \mu_z. \IEEEyessubnumber \label{eq:torsional_friction}
\end{IEEEeqnarray}
Being the constraints unilateral, condition \eqref{eq:pos_vert_forces} imposes that the force normal to the contact is greater than a value $f_z^{min}$, which must be greater than or equal to zero.
Eq. \eqref{eq:static_friction} limits the magnitude of the forces parallel to the contact surface not to overcome the static friction characterised by the coefficient $\mu_c$. Conditions \eqref{eq:CoP_y}-\eqref{eq:CoP_x} constrain the local Center of Pressure -- see, e.g.,~\cite[Appendix B, Eq. (A7)]{Frontiers2015} -- to remain inside the contact surface, which is assumed to be rectangular of dimensions $x_c^{min},x_c^{max},y_c^{min},y_c^{max}$, calculated w.r.t. the contact reference frame $\mathcal{C}$ and defined as shown in Fig. \ref{fig:foot_size}. Eq.~\eqref{eq:torsional_friction} imposes no foot rotation along the axis normal to the contact surface, and $\mu_z$ is the torsional friction coefficient. 

%% file: tex/statement.tex
\subsection{Problem Statement}
\label{sub-section:classical-qp}
\vspace{-0.05cm}

A common control approach for system~\eqref{eq:constrained_dynamics} usually considers several control objectives  organized  in  a  hierarchical or  weighted  prioritization \cite{mansard2009,nava2016}.
More precisely, let $a^*$ be a desired acceleration that the system should achieve. Then, a single priority\footnote{When several priorities are defined into the optimisation problem, higher priority tasks can be defined as  constraints of~\eqref{eq:sot}.}  stack-of-task can be represented by the following optimisation problem:
\vspace{-0.1cm}
\begin{IEEEeqnarray}{LLL}
  \label{eq:sot}
    \minimize_{y=(\dot{\nu},f,\tau)} ~&  \norm{Ay-a^*}^2 \\
    \text{subject to:}~& \nonumber 
    \\ &\begin{bmatrix} M(q) & -J^\top & -B \\ J & 0 & 0 \end{bmatrix}
         \begin{bmatrix}\dot{\nu} \\ f \\ \tau \end{bmatrix} {=} 
         \begin{bmatrix} -h(q,\nu) \\ -\dot{J}\nu \end{bmatrix} \nonumber \\
         &f  \in \mathcal{K} \nonumber
\end{IEEEeqnarray}
with $A$ a proper projection matrix, and $\mathcal{K}$ the manifold given by the constraints~\eqref{eq:contact_constraints}. The above optimisation problem is usually framed as a Quadratic Programming (QP) one, and its  solutions may suffer from the following limitations:
\begin{enumerate}
    \item The 
    solution may be discontinuous, e.g. at contact switching or after reference trajectory sharp variations;
    \item The closed-loop dynamics does not include any feedback term from the measured contact wrenches $f_{m}$. 
\end{enumerate}
Limitation $1)$ is often addressed by approximating the \emph{continuity property} with a set of inequality constraints to be added to~\eqref{eq:sot}, but the effectiveness of this approach is often unsatisfactory from the experimental standpoint~\cite{dafarra2018}. Limitation $2)$ is the most critical one, since FT sensor information are not used in the optimal control law $\tau$ that solves~\eqref{eq:sot}, thus potentially wasting important feedback information at the control level. 

Let us observe that Limitation $2)$ may be attenuated when desired force tasks are added to the problem~\eqref{eq:sot} \cite{bouyarmane2018}. For instance, if we aim to achieve  a desired force $f_d$, then the force task can be achieved by adding equality constraints in the form $f = f_d$ to the problem~\eqref{eq:sot}. At this point, one may attempt at using the FT measurements by replacing $f_d$ with 
\vspace{-0.11cm}
\[f^{*} = f_d -K_i\int_0^t (f_{m} - f_d)ds,
\vspace{-0.08cm}\]
where $f_m$ is the measured force, and $f = f^{*}$ being the equality constraint. The main limitation of this approach is that this equality constraint may require $f$ to violate the constraint $f  \in \mathcal{K}$. Putting the desired force as part of the cost function of~\eqref{eq:sot} may be an option, but this alters the priorities that the force task has over the acceleration one.

What follows presents an alternative, theoretically sound approach that aims at addressing the above limitations 
$1)$, and~$2)$ of classical QP based stack-of-task approaches for the control of floating base systems in contact with the environment.

%% file: tex/parametrization.tex
\section{A Contact-Stable Wrench Parametrization}
\label{sec:parametrization}

Parametrisations can be an effective way to transform constrained optimisation problems into unconstrained ones~\cite{marie2016}. Consider, for instance,  the following optimisation problem:
\begin{IEEEeqnarray}{LLL}
\IEEEyesnumber
  \label{eq:minConstraned}
      \minimize_{y} ~&  \text{cost}(y)  \\
\text{subject to}~& \nonumber \\ & y>0 \nonumber,
\end{IEEEeqnarray}
with $\text{cost}(\cdot):\mathbb{R}{\rightarrow}\mathbb{R}$, and $y\in\mathbb{R}$. If there exists a solution to~\eqref{eq:minConstraned}, the process of seeking for this solution is equivalent to solving the following problem:
\begin{IEEEeqnarray}{LLL}
\IEEEyesnumber
  \label{eq:minUnConstraned}
      \minimize_{\xi} ~&  \text{cost}(e^\xi),  
\end{IEEEeqnarray}
with $\xi \in \mathbb{R}$. 
For the specific case \eqref{eq:minConstraned}, it is trivial to find a parametrisation ensuring $y>0$.
Note, however, that the mapping $y = e^\xi$ is one-to-one, and its gradient is always invertible, namely $\frac{\partial}{\partial \xi}(e^\xi) = e^\xi \ne 0 \ \forall \xi$. These two additional properties are of particular importance for the numerical stability of solvers addressing the problem~\eqref{eq:minUnConstraned}.

Next Lemma proposes  a  wrench parametrisation that may be used to remove the constraint $f  \in \mathcal{K}$ into the 
problem~\eqref{eq:sot}. 
\newtheorem{lemma}{Lemma}
  \begin{lemma}
    \label{lemma_1}
    Consider the  parametrization  $f^k {=} \phi(\xi): \mathbb{R}^6 \rightarrow  \mathcal{K'}$ defined by
    \begin{equation}
      \label{eq:parametrized_wrench}
      \phi(\xi) :=
      \begin{bmatrix}
      \,{\mu_c}\frac{\tanh(\xi_1) \,(e^{\xi_3} + f_z^{min})}{\sqrt[]{1+\tanh^2(\xi_2)}}\\
      \,{\mu_c}\frac{\tanh(\xi_2) \,(e^{\xi_3} + f_z^{min})}{\sqrt[]{1+\tanh^2(\xi_1)}}\\
      e^{\xi_3} + f_z^{min}\ \\
      (\delta_y \tanh(\xi_4) + \delta_{y_0})\, (e^{\xi_3} + f_z^{min})\\
      (\delta_x \tanh(\xi_5) + \delta_{x_0})\, (e^{\xi_3} + f_z^{min})\\
      \mu_z \tanh(\xi_6)\, (e^{\xi_3} + f_z^{min})
      \end{bmatrix},
    \end{equation}
    with $f_z^{min} \geq 0$  the minimum value of the vertical force $f_z$,
    \begin{IEEEeqnarray}{LCL}
    \IEEEyesnumber
    \label{eq:feet_size_constraints}
    \delta_x := 
    \frac{x_c^{max}- x_c^{min}}{2}, \quad
    \delta_{x_0} := -
      \frac{x_c^{min}+x_c^{max}}{2} \IEEEyessubnumber \\
      \delta_y := 
      \frac{y_c^{max} - y_c^{min}}{2}, \quad \delta_{y_0} := 
      \frac{y_c^{max} + y_c^{min}}{2},  \IEEEyessubnumber
%
    \end{IEEEeqnarray}
    and $x_c^{max}, y_c^{max}$ and $x_c^{min}, y_c^{min}$  the contact surface dimensions as described in Fig. \ref{fig:foot_size}. 
     
    \noindent
    Then, the following properties hold:
    \begin{enumerate}
      \item The contact constraints \eqref{eq:contact_constraints} are always satisfied, namely, $\mathcal{K'} \subset \mathcal{K}$ or, equivalently, $\phi(\xi) \in \mathcal{K} \ \ \forall \xi \in \mathbb{R}^6 $.
     \item The function $\phi(\xi): \mathbb{R}^6 \rightarrow  \mathcal{K'}$ is a bijection, namely, a one-to-one correspondence from $\mathbb{R}^6 $ to $ \mathcal{K'}$.
      \item 
      The gradient of the function $\phi(\xi)$, i.e. 
      \begin{IEEEeqnarray}{LLL}
      \label{gradient-phi}
      \Phi(\xi) &:= \left[\frac{\partial\phi}{\partial \xi_1}, ... \hspace{0.5 mm}, \frac{\partial\phi}{\partial \xi_6}\right]\in \mathbb{R}^{6 \times 6},
      \end{IEEEeqnarray}
      is an invertible matrix $\forall \hspace{0.5 mm}\xi \in \mathbb{R}^6$.
    \end{enumerate}
\end{lemma}
The proof is in the Appendix \ref{appendix_1}. Lemma~\ref{lemma_1} shows that  \eqref{eq:parametrized_wrench} generates wrenches that always belong to the contact stability manifold defined by \eqref{eq:contact_constraints}. Furthermore, it shows that $\phi(\cdot)$ is a one-to-one correspondence between a set of free parameters~$\xi$ and the manifold $\mathcal{K'}$, i.e. the 
image of \eqref{eq:parametrized_wrench}. Clearly, one may find other functions for which the contact constraints~\eqref{eq:contact_constraints} are always satisfied. The proposed function $\phi     (\cdot)$ in~\eqref{eq:parametrized_wrench}, however, has an image that 
approximates the friction cones \eqref{eq:static_friction} similarly to 
what one obtains using a set of linear inequalities.
%
\begin{figure}[ht!]
  \centering
  \begin{subfigure}[b]{0.3\textwidth}
    \includegraphics[width=\textwidth]{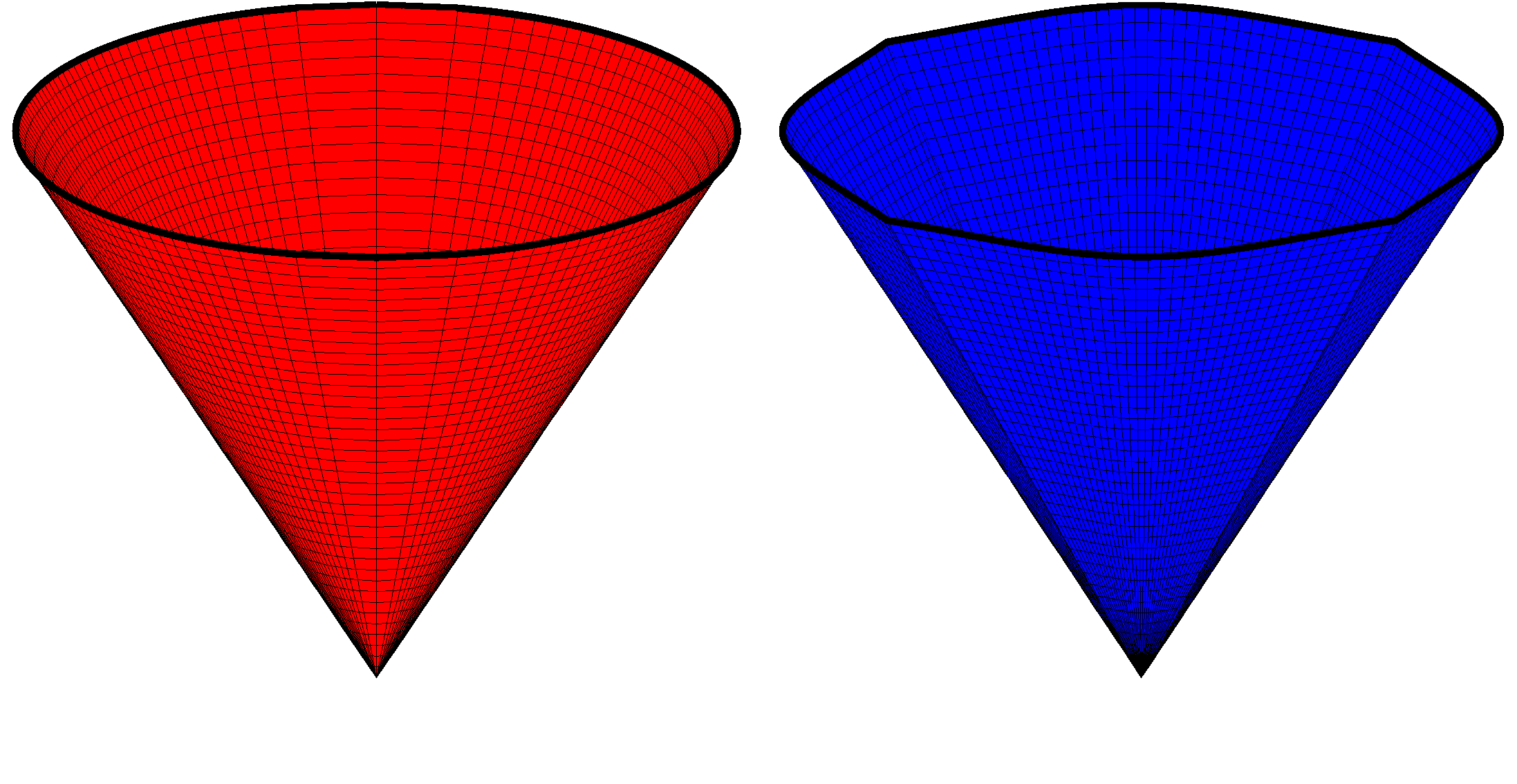}
  \end{subfigure}
  \begin{subfigure}[b]{0.18\textwidth}
    \includegraphics[width=\textwidth]{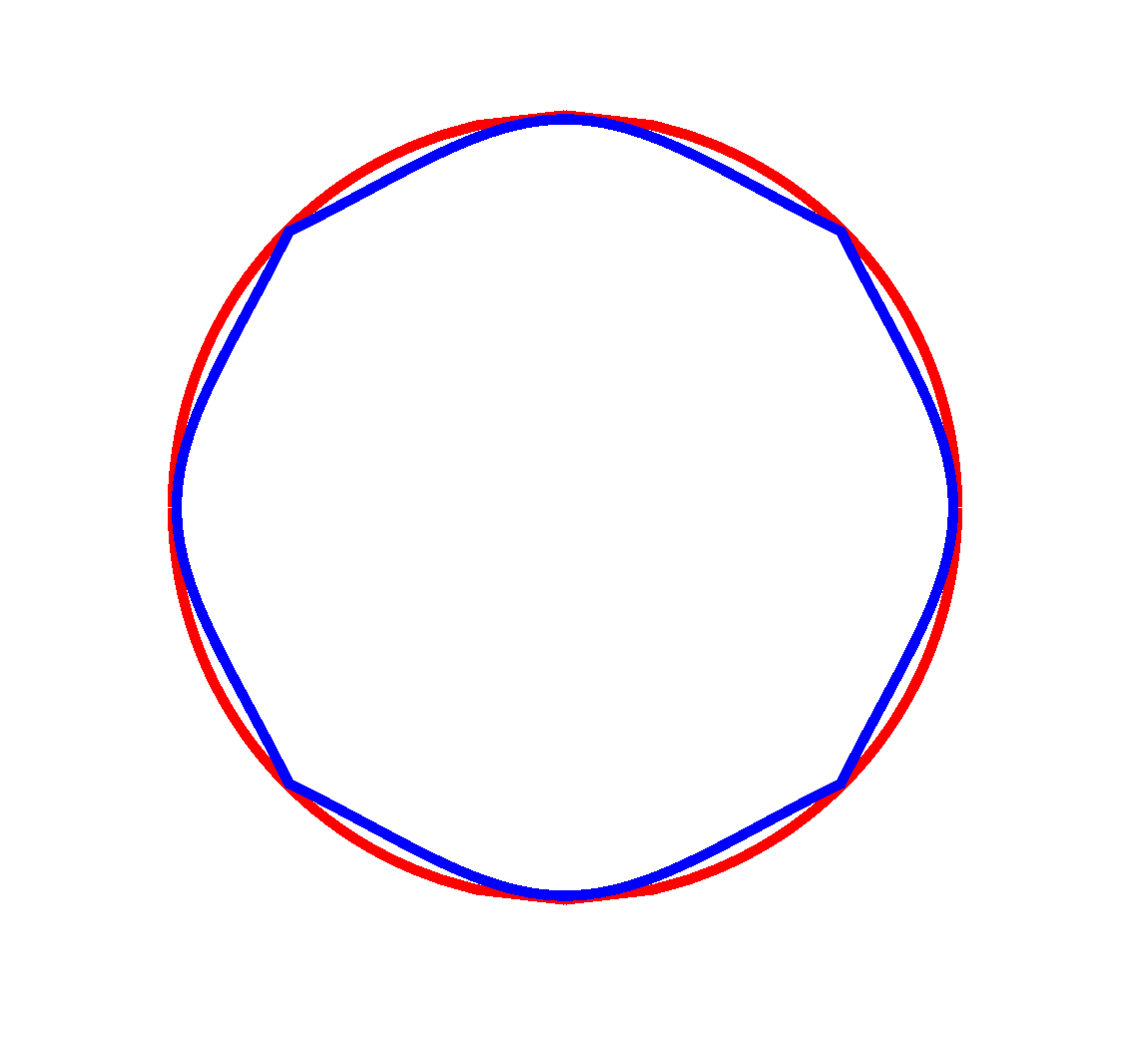}
  \end{subfigure}
  \caption{A comparison between the friction cone (left) and its parametrisation with hyperbolic tangents (right). On the right: top view of the two manifolds.}
  \label{figure_approximation_params}
\end{figure}
More precisely, 
when \eqref{eq:parametrized_wrench} is substituted into
\eqref{eq:static_friction}, instead of a cone we obtain a set that much resembles an octagonal flipped pyramid. In fact, it is easy to verify that $\sqrt[]{f_x^2 + f_y^2} = \mu_c f_z$ is satisfied 
eight times when $f^k {=} \phi(\xi)$ and either, or both, $\tanh^2(\cdot) = 1$. By computing the ratio between the areas of an octagon and that of a circle, we conclude that \eqref{eq:parametrized_wrench} covers more than 90 \% of the set given by \eqref{eq:static_friction}.
For a visual representation of this fact, see Figure~\ref{figure_approximation_params} that shows a typical approximation of \eqref{eq:static_friction} given by the function~\eqref{eq:parametrized_wrench}. 
Let us also observe that the gradient of this function is invertible for any value of the parameter~$\xi$. This property will be of pivotal importance in Sections~\ref{sec:control} and \ref{sec-jerk-momentum-control} when designing  stable controllers for system~\eqref{eq:constrained_dynamics}.

The  parametrization~\eqref{eq:parametrized_wrench} can be easily extended in case of $n_c$ distinct contact wrenches. In this case, define:
\begin{IEEEeqnarray}{LCL}
\IEEEyesnumber
  \label{eq:parametrization_multiple_contacts}
  f &=& [f^{1}; ... \hspace{0.5 mm}; f^{n_c}] := [\phi({\xi^{1}}); ... \hspace{0.5mm}; \phi({\xi^{n_c}})], \IEEEyessubnumber \\
  \dot{f} &=& \Phi(\xi)\dot{\xi}, \IEEEyessubnumber
\end{IEEEeqnarray}
where $\Phi = \text{blkdiag}(\Phi_1, ... \hspace{0.5 mm}, \Phi_{n_c}) \in \mathbb{R}^{6n_c \times 6n_c}$ and $\xi = [\xi^1; ... \hspace{0.5 mm}; \xi^{n_c}] \in \mathbb{R}^{6n_c}$. It is then straightforward to verify that the properties described in Lemma \ref{lemma_1} are retained even in case of multiple contact wrenches.

%% file: tex/control.tex
\section{Jerk Control}
\label{sec:control}

This section proposes control laws that exploit the contact wrench parametrisation~\eqref{eq:parametrized_wrench} and attempt to address the limitations -- listed in Section~\ref{sub-section:classical-qp}  --  of the classical torque-based controllers framed as stack-of-tasks optimisation problems. 

\subsection{
Jerk control with parametrised contact wrenches}
The wrench parametrisation~\eqref{eq:parametrized_wrench} can be used to remove the constraint $f  \in \mathcal{K}$ from the optimisation problem~\eqref{eq:sot}. This process would lead to the following formulation:
\begin{IEEEeqnarray}{LLL}
\IEEEyesnumber
  \label{eq:sot-noK}
    \minimize_{y=(\dot{\nu},\xi,\tau)} ~&  \norm{Ag(y)-a^*}^2 \\
    \text{subject to:}~& \nonumber 
    \\ &\begin{bmatrix} M(q) & -J^\top & -B \\ J & 0_{6n_c} & 0_{6n_c,n} \end{bmatrix}
         \begin{bmatrix}\dot{\nu} \\ \phi(\xi) \\ \tau \end{bmatrix} {=} 
         \begin{bmatrix} -h(q,\nu) \\ -\dot{J}\nu \end{bmatrix} \nonumber 
\end{IEEEeqnarray}
with $g(y):= [\dot{\nu};\phi(\xi);\tau]$. The main drawbacks of the above approach are: $i)$ the optimisation problem~\eqref{eq:sot-noK} can no longer be casted in a QP being the parametrisation $\phi(\xi)$ nonlinear; we would then need nonlinear -- and often slower than QPs -- optimisers to solve~\eqref{eq:sot-noK};  $ii)$ the limitations 1), 2) listed in Section~\ref{sub-section:classical-qp} are not addressed. 

To include feedback terms into the control laws, the contact wrenches, or accelerations, shall become part of the system state. In the language of Automatic Control, we shall then proceed with augmenting the relative degree of the output (or task) that one wants to stabilise~\cite{isidori2013}.

More precisely, assume that: $hp{-}i)$ the control objective is the stabilisation of a desired jerk $\dot{a}^*$; $hp{-}ii)$  the joint torque rate-of-change $\dot{\tau}$ can be considered as a control input; $hp{-}iii)$ both the joint torques $\tau$ and the contact forces $f$ are measurable quantities, so the robot acceleration $\dot{\nu}$ -- if not measurable -- can be obtained from~\eqref{eq:constrained_dynamics}. Now, define 

\begin{IEEEeqnarray}{RCL}
\IEEEyesnumber
    D&:= &\begin{bmatrix} M(q) & -J^\top & -B \\ J & 0_{6n_c} & 0_{6n_c,n} \end{bmatrix}, \IEEEyessubnumber \\
    \beta &:=& \begin{bmatrix} -h(q,\nu) \\ -\dot{J}\nu \end{bmatrix}.  \IEEEyessubnumber
\end{IEEEeqnarray}
As a consequence of $hp{-}iii)$, one has a measurement of the vector $y=(\dot{\nu},f,\tau)$, while the variable $\dot{y}$ can be used as a search variable. Then, control laws for $\dot{\tau}$ that contain feedback information from the FT sensors are obtained as an outcome of the following optimisation problem:
\begin{IEEEeqnarray}{LLL}
    \IEEEyesnumber
    \label{eq:sotJerk}
    \minimize_{\dot{y}=(\ddot{\nu},\dot{f},\dot{\tau})} ~&  \norm{\dot{A}y + A\dot{y} -\dot{a}^*}^2 \IEEEyessubnumber \\
    \text{subject to:}~& \nonumber 
    \\ &\dot{D}y+D\dot{y} = \dot{\beta}\nonumber  \\
         &  f  \in \mathcal{K}.    \IEEEyessubnumber \label{eq:finKsotJerk}
\end{IEEEeqnarray}
The solutions to the above problem are continuous in $\tau$ (even if $\dot{\tau}$ is discontinuous) and contain FT measurement feedback from the vector $y$. One of the main difficulties when solving~\eqref{eq:sotJerk} is given by~\eqref{eq:finKsotJerk}. Since the variable $f$ no  longer is a search variable, in fact, one cannot instantaneously choose values of the contact wrenches such that $f  \in \mathcal{K}$. One  may attempt to make~\eqref{eq:finKsotJerk} satisfied by regulating appropriately the variable~$\dot{f}$, which influences the wrench $f$ at the next time step. 

A possibility to make~\eqref{eq:finKsotJerk} satisfied is to use the parametrisation in Lemma~\ref{lemma_1}: the gradient of the parametrisation automatically enforces the fact that $f(t)  \in \mathcal{K} \ \forall t$. 
More precisely, in view of~\eqref{gradient-phi}, one has $\dot{y} = (\ddot{\nu},\Phi(\xi)\dot{\xi},\dot{\tau})$, which leads to the following optimisation problem:
\begin{IEEEeqnarray}{LLL}
\IEEEyesnumber
    \label{eq:sotJerkNoK}
    \minimize_{u=(\ddot{\nu},\dot{\xi},\dot{\tau})} ~&  \norm{\dot{A}y + APu -\dot{a}^*}^2 \IEEEyessubnumber  \\
    \text{subject to:}~& \nonumber 
    \\ &\dot{D}y+DPu = \dot{\beta}, \IEEEyessubnumber  
\end{IEEEeqnarray}
with $P$ defined as: 
\[
P:= 
\begin{pmatrix}
\mathbf{I}_{n+6} & 0_{n+6,6n_c} & 0_{n+6,n} \\ 
0_{6n_c,n+6} & \Phi(\xi) & 0_{6n_c,n} \\
0_{n,n+6} & 0_{n,6n_c} & \mathbf{I}_n
\end{pmatrix}.
\]
In order to be solved at each time instant, the optimisation problem~\eqref{eq:sotJerkNoK} requires the variable $\xi$. This variable may be retrieved from either time integration of~$\dot{\xi}$, or by inverting the relationship $f = \phi(\xi)$:  being the parametrisation a one-to-one correspondence (see Lemma~\ref{lemma_1}), there exists a unique $\xi$ for any value of the contact wrench $f$ provided that it belongs to $\mathcal{K'} \subset \mathcal{K}$.
The latter allows us to inject further information from the FT sensor measurements into the optimal control laws $u$ solving the optimisation problem~\eqref{eq:sotJerkNoK}.

Note also that the matrix $P$ is invertible thanks to the property $3)$ of Lemma~\ref{lemma_1}. The invertibility of $P$ clearly plays
a pivotal role when solving the optimisation problem~\eqref{eq:sotJerkNoK}.

\subsection{On the modeling and control requirements for jerk control} 
\label{sub-tauDot}
The optimal value $\dot{\tau}$ solving~\eqref{eq:sotJerkNoK} may be sent directly to the robot low-level control system if it allows to set desired rate-of-changes of joint torques. This may be feasible, for instance, when the low-level robot control  exploits the model between the joint torques $\tau$ and the motor currents $i$, e.g. $\tau = k_\tau i$. More precisely, the motor currents are usually subject to electrical dynamics of the kind $\tfrac{d}{dt}{i} = k_i v$, where $v$ is often the motor voltages to be applied to the motors -- namely, the real control input. Then, it is straightforward to express the optimisation problem~\eqref{eq:sotJerkNoK} so that the search variable $u$ contains~$v$. Let us observe, however, that this control architecture in general requires high-frequency control loops (e.g. $5-20$ KHz) for generating the motor voltages~$v$: these loops have to compute inverse dynamics within a short control cycle. If the control loops are not fast enough, sampling effects may be preponderant phenomena that render the assumption $\tfrac{d}{dt}{i}~=~k_i v$ not representative of the underlying physical dynamics. In this case, the associated control strategy resulting from~\eqref{eq:sotJerkNoK} may prove to be ineffective.

Another necessary requirement for achieving jerk control is the calculation of the terms $\dot{A}$, $\dot{D}$ and $\dot{\beta}$ to solve the optimisation problem~\eqref{eq:sotJerkNoK}. These terms in general depend on the robot configuration space $q$, velocity $\nu$, and accelerations $\ddot{\nu}$, and need the derivatives of the system inverse dynamics. Besides numerical approximations for computing these terms, existing libraries nowadays provide users with the support of automatic differentiation and/or directly derivatives of inverse dynamics~\cite{carpentier2018analytical,Andersson2018}. If some of the terms in~\eqref{eq:sotJerkNoK} are not available, one may attempt setting them equal to zero and 
tune the feedback control gains in $\dot{a}^\ast$ so as to achieve robustness against them. 
However, we present below a jerk control architecture  
that overcomes the above  modeling and control limitations of the mere application of~\eqref{eq:sotJerkNoK}.

\section{Momentum-based Jerk control}
\label{sec-jerk-momentum-control}

This section proposes control laws that can be obtained from the problem~\eqref{eq:sotJerkNoK} when it is explicitly solved  and extended  for a two layer stack-of-task. These laws can also be shown to possess  stability properties. Interestingly, the architecture presented below does not need the feedforward terms that depend on the inverse dynamics derivatives required by~\eqref{eq:sotJerkNoK}. This is achieved by loosing the continuity property of $\tau$ but retaining the continuity of the contact wrenches $f$.  

More precisely, we assume that: i) the highest priority task is the  stabilisation of a desired robot  centroidal momentum~\cite{traversaro2017,orin08}; ii) the lower priority task aims at stabilising the robot \emph{posture} to regulate the system \emph{zero dynamics} \cite{isidori2013}. 

Let us recall that the momentum rate-of-change equals the summation of all the external wrenches acting on the robot. In a multi-contact scenario, the external wrenches reduce to the contact wrenches plus the gravity force:
\begin{IEEEeqnarray}{LCL}
\IEEEyesnumber
  \dot{H} &=& \sum_{k = 1}^{n_c} A_kf^k - mge_3 = Af -mge_3, \IEEEyessubnumber \label{eq:centroidal_momentum}  \\
  A &:=& \hspace{0.5 mm} [A_1, ... \hspace{0.5 mm}, A_{n_c}] \in \mathbb{R}^{6 \times 6n_c},   \IEEEyessubnumber \label{matrix_A} \\ %
  A_k &=& \begin{bmatrix}
  1_3  &  0_{3} \\
  S(\,^{\mathcal{I}}o_{\mathcal{C}_k} - \,^{\mathcal{I}}o_{CoM}) & 1_3 \end{bmatrix},  \IEEEyessubnumber \label{matrix_Ak}
\end{IEEEeqnarray}
where $H \in \mathbb{R}^6$ is the robot's momentum, $A_k \in \mathbb{R}^{6 \times 6}$ is the matrix mapping the $k$-th contact wrench to the momentum dynamics, $^{\mathcal{I}}o_{\mathcal{C}_k} \in \mathbb{R}^3$ is the origin of the frame associated with the $k$-th contact, and $^{\mathcal{I}}o_{CoM} \in \mathbb{R}^3$ is the CoM position. 

Recall that the rate-of-change of the robot momentum~\eqref{eq:centroidal_momentum} is related to the system accelerations (e.g. acceleration of the system center of mass). So, to derive jerk-based control laws, we need to differentiate~\eqref{eq:centroidal_momentum} w.r.t. time, which writes:
\begin{IEEEeqnarray}{LCL}
\IEEEyesnumber
  \label{eq:centroidal_momentum_acc}
  \ddot{H} &=& A\dot{f} + \dot{A}f = A\Phi(\xi)\dot{\xi} + \dot{A}f, \\
  \dot{A}  &:=& \hspace{0.5 mm} [\dot{A}_1, ... \hspace{0.5 mm}, \dot{A}_{k}]\hspace{0.5 mm} \hspace{0.5 mm} \forall\hspace{0.5 mm}{k} = 1, ... \hspace{0.5 mm}, n_c, \nonumber \\
  \dot{A}_k &=& \begin{bmatrix}
                  0_3  &  0_{3} \\
                  S(\,^{\mathcal{I}}\dot{o}_{\mathcal{C}_k} - \,^{\mathcal{I}}\dot{o}_{CoM}) & 0_3 \end{bmatrix}. \nonumber
\end{IEEEeqnarray}
Note that Eq. \eqref{eq:centroidal_momentum_acc} is linear w.r.t. $\dot{\xi}$ and optimisation problems similar to~\eqref{eq:sotJerkNoK} may be laid down. In particular, to obtain the robot momentum stability, one may: $i)$~consider $\dot{\xi}$ as control input -- or search variable -- of the momentum acceleration~\eqref{eq:centroidal_momentum_acc};  $ii)$~apply feedback linearization to~\eqref{eq:centroidal_momentum_acc} in order to impose a momentum acceleration $\ddot{H}^*$ of the form:
\begin{IEEEeqnarray}{LCL}
  \label{eq:closed_loop_fb_lin}
  \ddot{H}^* = \ddot{H}_d - K_d\dot{\tilde{H}} - K_p{\tilde{H}}, 
\end{IEEEeqnarray}
where $K_d,K_p \in \mathbb{R}^{6 \times 6}$ are symmetric and positive definite matrices, $H_d\in \mathbb{R}^{6}$ is the reference momentum, and $\tilde{H} = H - H_d$ is the momentum error. 
%
Observe that it is always possible to find~$\dot{\xi}$ such that 
\begin{IEEEeqnarray}{LCL}
  \label{eq:closed_loop_fb_linH}
  \ddot{H}(\dot{\xi}) = \ddot{H}^*
\end{IEEEeqnarray}
because of the item $3)$ of Lemma~\ref{lemma_1}. More precisely, the gradient~$\Phi$ being always invertible ensures that the matrix~$A\Phi$ in~\eqref{eq:centroidal_momentum_acc} is full rank $\forall \ \xi$. Consequently, $\dot{\xi}$ has full control authority on the momentum acceleration for any value of $\xi$. Clearly, one can impose~\eqref{eq:closed_loop_fb_linH} as long as $\xi$ remains bounded.

The stability properties of the control laws ensuring \eqref{eq:closed_loop_fb_linH} are then presented below.

\newtheorem{theorem}{Theorem}
  \begin{lemma} 
  \label{lemma_12}
  Assume that: 
  \begin{itemize}
      \item the robot makes at least one rigid contact with the environment, i.e. $n_c \geq 1$;
      \item the desired momentum $H_d$ is a feasible system equilibrium such that there exists $f_e(t)  \in \mathcal{K'} $ satisfying \[\dot{H}_d = Af_e -mge_3;\]
      \item the variable $\dot{\xi}$ is chosen so as \eqref{eq:closed_loop_fb_linH} is satisfied, with $\ddot{H}$ given by \eqref{eq:centroidal_momentum_acc} and $\ddot{H}^*$  by \eqref{eq:closed_loop_fb_lin}.
  \end{itemize}
  Then:
  \begin{enumerate}
      \item the equilibrium point $(\tilde{H},\dot{\tilde{H}}) = (0,0)$ is locally asymptotically stable if the robot makes one rigid contact with the environment, i.e. $n_c = 1$;
      \item the equilibrium point $(\tilde{H},\dot{\tilde{H}}) = (0,0)$ is globally asymptotically stable if $\xi$ is bounded, namely if there exists a constant $c \in \mathbb{R}^+$ such that $|\xi(t)| < c \ \forall t$.   
  \end{enumerate}
  \end{lemma}
The proof is in the Appendix \ref{appendix_12}. Lemma \ref{lemma_12} shows that as long as we satisfy Eq. \eqref{eq:closed_loop_fb_linH}, the system trajectories converge towards the desired values. The possibility of satisfying Eq. \eqref{eq:closed_loop_fb_linH} is inherently related to the boundedness of $\xi$, which guarantees that the matrix $A\Phi(\xi)$ in \eqref{eq:centroidal_momentum_acc} remains of full rank. More precisely, the item $1)$ shows that the boundeness of $\xi$ is achieved locally to the equilibrium point when the number of contacts is equal to one. When the number of contacts is greater than one,  there is a redundancy for $\dot{\xi}$ that solve \eqref{eq:closed_loop_fb_linH}, and this redundancy should be chosen so as $\xi$ is always bounded. In this case, in fact, the item $2)$ of Lemma \ref{lemma_12} shows that the system trajectories globally converge towards the desired values. We show in the next section a possible choice for the redundancy of $\dot{\xi}$ that proved to work effectively both in simulation and in real experiments. Let us remark, however, that the aim of a global bounded $\xi$ cannot be in general achieved, and so global asymptotic stability. For instance, think of a humanoid robot standing on one foot and starting with an high velocity of its center of mass: there is a limit for this velocity that would cause the contact to break, namely, the variable $\xi$ growing indefinitely. However, we can monitor the adverse conditions, where contacts are about to break, by looking at the overall norm of the variable $\xi$. This is, in the authors' opinion, an interesting result of the proposed approach. 

Often, the control objective of the center-of-mass trajectory tracking is framed as momentum control. In this case, one is tempted to use the approach presented above with 
\begin{IEEEeqnarray}{LCL}
  \label{eq:closed_loop_fb_lin_int}
  \ddot{H}^* = \ddot{H}_d - K_d\dot{\tilde{H}} - K_p{\tilde{H}} - K_i\int_0^t{\tilde{H}} dt, \\ 
  K_i = K_i^\top > 0, \nonumber
\end{IEEEeqnarray}
where the (linear momentum) integral correction terms can be replaced by the position errors between the center of mass trajectory and its desired values. 
The resulting third order system~\eqref{eq:closed_loop_fb_linH}-\eqref{eq:closed_loop_fb_lin_int}, however, is in general very sensitive to gain tuning, as not all  possible combinations of the gain matrices guarantee stability of the associated closed-loop  system. This limitation affects the controller's performances when applied to the real robot, where phenomena as  modeling errors, measurements noise and external disturbances further limit the control gain choice.

\subsection{Momentum-based jerk control with integral terms}

We propose a control algorithm alternative to \emph{pure} feedback linearization with the goal of facilitating the gain tuning of the closed-loop system dynamics. In particular, consider as control objective the stabilization of $(I,\tilde{H},\zeta)$ 
towards the reference values $(0,0,0)$, with $I$ the integral of the momentum error, $\tilde{H}$ the momentum error, and $\zeta$ an exogenous state that will be used to  prove  Lyapunov  stability, i.e.
\begin{IEEEeqnarray}{LCL}
\IEEEyesnumber
  \label{eq:output_ext}
  I &:=& \int_0^t\tilde{H}dt,  \IEEEyessubnumber
  \label{eq:I}\\ 
  \tilde{H}&:=& H - {H}_d,    \IEEEyessubnumber
  \label{eq:tilde_H}\\ 
  \zeta &:=& Af -mge_3 - \dot{H}_d + K_d \tilde{H} + K_p I,  \IEEEyessubnumber 
  \label{eq:zeta}
\end{IEEEeqnarray}
whose dynamics write:
\begin{IEEEeqnarray}{LCL}
\IEEEyesnumber
  \label{eq:output_dyn}
  \dot{I} &:=& \tilde{H},  \IEEEyessubnumber \label{eq:dot_I}\\ 
  \dot{\tilde{H}} &:=& Af -mge_3 -\dot{H}_d = \zeta - K_d \tilde{H} - K_p I, \IEEEyessubnumber \label{eq:dot_tilde_H}\\
  \dot{\zeta} &:=& \dot{A}f + A\Phi(\xi)\dot{\xi} - \ddot{H}_d + K_d \dot{\tilde{H}} + K_p\tilde{H}.  \IEEEyessubnumber \label{eq:dot_zeta}
\end{IEEEeqnarray}
 The dynamics \eqref{eq:output_dyn} are obtained by taking the derivative of $(I,\tilde{H},\zeta)$, and 
 substituting their corresponding definitions  in (\ref{eq:I})-(\ref{eq:zeta}). Then, the following result holds.
 
  \begin{lemma} 
  \label{lemma_2}
  Assume that: 
  \begin{itemize}
      \item the robot makes at least one rigid contact with the environment, i.e. $n_c \geq 1$;
      \item the desired momentum $H_d$ is a feasible system equilibrium such that there exists $f_e(t)  \in \mathcal{K'} $ satisfying \[\dot{H}_d = Af_e -mge_3.\]
  \end{itemize}
  Choose,
  \begin{align} 
    \label{eq:input_momentum}
    \dot{\xi} = \, & {(A\Phi)}^{\dagger}\,[\ddot{H}_d - (K_d + 1_6)\dot{\tilde{H}} \\
                     & - (K_d + K^{-1}_o + K_p)\tilde{H} - K_p I - \dot{A}f] \nonumber \\
                      &+ N_{A\Phi}\dot{\xi}_0 \nonumber, 
  \end{align}
  with $K_o,K_p,K_d {\in}  \mathbb{R}^{6 \times 6}$ symmetric positive definite matrices, \[N_{A\Phi} = (1_6 - {(A\Phi)}^{\dagger}A\Phi)\] the projector in the null space of $A\Phi$, and $\dot{\xi}_0$ a free variable of proper dimension. Then:
  \begin{enumerate}
      \item the equilibrium point $(I,\tilde{H},\zeta) = (0,0,0)$ is locally asymptotically stable if the robot makes one rigid contact with the environment, i.e. $n_c = 1$;
      \item the equilibrium point $(I,\tilde{H},\zeta) = (0,0,0)$ is globally asymptotically stable if $\xi$ is bounded, namely if there exists a constant $c \in \mathbb{R}^+$ such that $|\xi(t)| < c \ \forall t$.  
  \end{enumerate}
  \end{lemma}
  The proof is in the Appendix \ref{appendix_2}. 
  Lemma~\ref{lemma_2} shows that the additional integral terms do not break the stability properties achieved in Lemma \ref{lemma_12}, and that one is left with free positive definite gains $K_o, \ K_p, \ K_d$. 
  Let us recall that the constraints~\eqref{eq:contact_constraints}  remain satisfied while ensuring the stability properties of the associated closed-loop system, and such a claim cannot usually be made in classical stack-of-task approaches~\eqref{eq:sot}. 

The control law~\eqref{eq:input_momentum} contains both feedforward and feedback terms that depend on the measured contact wrenches. It makes use, in fact, of~\eqref{eq:centroidal_momentum} for computing $\dot{H}$, which depends on the measured contact wrenches.
In case of a single contact, there exists a unique control input $\dot{\xi}$ that satisfies~\eqref{eq:input_momentum}, and the null space of the matrix $A\Phi$ is empty, i.e. $N_{A\Phi} = 0$. In case of multiple contacts ($n_c > 1$), instead, infinite control inputs satisfy~\eqref{eq:input_momentum}. We solve the associated redundancy using the free variable $\dot{\xi}_0$ to minimize the norm of the robot joint torques. The computation of $\dot{\xi}_0$ is detailed in Appendix \ref{computation-xi-0}.

Let us remark again the importance of the invertibility of the gradient $\Phi$ -- see Lemma~\ref{lemma_1}. This property guarantees that the matrix  $A\Phi$ in~\eqref{eq:input_momentum} is full rank, so $\dot{\xi}$ has full control authority on the momentum acceleration for any value of $\xi$.

\subsection{Computation of $f$, $\dot{H}(f)$ and $\Phi(\xi)$} 
\label{subsec:get_dyn_params}

The control input~\eqref{eq:input_momentum} requires: the contact wrenches~$f$; the momentum derivative $\dot{H} = \dot{H}(f)$; and the associated variable $\xi$ such that $f = \phi(\xi)$. The contact wrenches can be measured/estimated using the measurements from 6-axis FT sensors installed on the robot. Once the wrenches $f$ are retrieved, we can compute the momentum rate of change via~\eqref{eq:centroidal_momentum}
The associated  $\xi$ can be computed by applying the parametrisation \emph{inverse} mapping, namely~$\xi = \phi^{-1}(f)$. The inverse mapping exists provided that the measured contact wrenches remain inside the set $\mathcal{K'}$. 
If the measured wrenches do not belong to $\mathcal{K'}$ (because, e.g., measurements noise, external unmodeled disturbances, etc.), a saturation shall be applied in the calculation of the inverse mapping so that the control input ${\xi}$ always remains finite.

\subsection{Computation of the joint torques to realize $\dot{\xi}$} 
\label{sub-xiDotControl}

To realize a $\dot{\xi}$, e.g. the law in~\eqref{eq:input_momentum},  we have to choose the \emph{real} control input of the system properly. We assume in this section that the control input is the joint torque $\tau$, so we cannot impose a desired $\dot{\tau}$ instantaneously.

A possibility for computing the joint torques is to find $\dot{\tau}$ realising $\dot{\xi}$, and then perform time-integration of $\dot{\tau}$ to obtain $\tau$. This procedure, however, requires to compute some derivatives of the inverse dynamics, which may not be available in practice.

For this reason, we follow here another route for finding the joint torques $\tau$ attempting to realise $\dot{\xi}$. First, we find the instantaneous relationship between the joint torques $\tau$ and the contact wrenches $f$. This relationship can be found, for instance, by substituting the state accelerations $\dot{\nu}$ from~\eqref{eq:dynamics_constr} into the constraints~\eqref{eq:constraint_acc}, which leads to:
\begin{IEEEeqnarray}{LCL}
  \label{eq:torques_forces}
  JM^{-1}(J^\top f -h) + \Lambda\tau + \dot{J}\nu = 0,
\end{IEEEeqnarray}
with $\Lambda = JM^{-1}B$. Then, we proceed as follows:
\begin{itemize}
    \item Integrate the control input $\dot{\xi}$ to obtain $\xi$. The initial conditions for the integrator can be calculated by measuring the initial contact forces 
    $f(0)$ 
    and by applying the parametrization \emph{inverse mapping}, i.e. 
    $\xi(0) = \phi^{-1}(f(0))$;
    \item Apply the parametrization direct mapping to evaluate the wrenches $f$ from $\xi$, i.e. $f = \phi(\xi)$. By doing so, note that $f$ always satisfy the contact stability constraints;
    \item Retrieve the input torques $\tau$ from~\eqref{eq:torques_forces}, which write
\end{itemize}
\begin{IEEEeqnarray}{LCL}
  \label{eq:input_torques}
  \tau = \Lambda^{\dagger}(JM^{-1}(h - J^\top f) - \dot{J}\nu) + N_{\Lambda}\tau_0,
\end{IEEEeqnarray}
where $N_{\Lambda} = (1_n - {\Lambda}^{\dagger}\Lambda)$ is the projector in the null space of $\Lambda$, and $\tau_0$ is a free variable, that can be chosen as in  \cite{nava2016} to guarantee the stability of the system's zero-dynamics.
%

%% file: tex/results.tex
\section{Simulations and Experimental Results}
\label{sec:results}

\subsection{Simulation Environment}

The modeling and control framework presented in Sec. \ref{sec-jerk-momentum-control} is tested on the 23-DoFs iCub humanoid robot \cite{Metta20101125}, both on the real robot and in simulations using Gazebo  \cite{Koenig04}. The controller is implemented in Simulink, and runs at a frequency of $100~\text{[Hz]}$. An advantage of using the Simulink-Gazebo simulator consists in the possibility to test directly on the real robot the same control software used in simulation. 

Gazebo offers different physic engines to integrate the system's dynamics. We chose the Open Dynamics Engine (ODE), that uses a fixed step semi-implicit Euler integration scheme, with a maximum simulation time step of $1$ [ms].

On the real iCub, the Simulink controller runs on an external PC and provides reference joint torques to an \emph{inner} joint torque control loop, that runs 
on board the robot at  $1000~\text{[Hz]}$. 
At the moment, iCub is not endowed with joint torque sensors. The measured joint torques are achieved by combining the FT sensor information, the joint encoders, IMU data and the robot model. 
%
%
The robot is equipped with 6 FT sensors: two located in the robot's upper arms, two in the robot thighs, and two in the robot's feet~\cite{Metta20101125}. During experiments on the real robot, we verified that the desired controller step time of $0.01 \ \rm [s]$ is respected most of the times. More precisely, statistics show that the desired step time is respected around $97.9 \%$ of iterations for the jerk control. This is comparable with the performances of the classical momentum-based QP control \cite{nava2016}, which meets the desired step time around $98.2 \%$ of iterations.

The choice of running the high-level Simulink controller with a frequency of  $100~\text{[Hz]}$ is due to: a CPU limitation of the current iCub; the limited bandwidth of the current FT sensors at the robot's feet. However, we are working on robot hardware upgrades that will allow us to run the Simulink controllers with a standard frequency of $1000~\text{[Hz]}$.

\subsection{Robustness analysis for jerk control}
\label{subsec:regularization}

A preliminary analysis of the robot balancing behavior with the control law \eqref{eq:input_momentum}--\eqref{eq:input_torques} indicates that the proposed jerk control strategy may be particularly sensitive to bias errors on the estimated momentum rate of change $\dot{H}$, which is used as feedback in Eq. \eqref{eq:input_momentum}. The momentum rate of change is estimated as detailed in Sec. \ref{subsec:get_dyn_params}. The sources of this bias may be errors in the robot dynamic parameters, or low FT sensor accuracy. To enforce the closed loop system robustness w.r.t. errors in the estimation of $\dot{H}$,
we modified Eq. \eqref{eq:input_momentum} by adding a regularization term as follows:
\begin{align} 
    \label{eq:input_momentum_modif}
    \dot{\xi} = \, & {(A\Phi)}^{\dagger}\,[\ddot{H}_d - (K_d + 1_6)\dot{\tilde{H}} \\
                     & - (K_d + K^{-1}_o + K_p)\tilde{H} - K_p I - \dot{A}f] \nonumber \\
                     & + N_{A\Phi}\dot{\xi}_0  -k_e(\xi -\xi_d)\nonumber, 
\end{align}
where $k_e > 0$ is a positive scalar, $\xi$ the integral of $\dot{\xi}$ and $\xi_d$ is obtained by applying the parametrization inverse mapping on the set of wrenches satisfying the desired momentum rate of change, i.e. $\dot{H}_d(f_d)$. In case of multiple solutions, the one ensuring minimum norm of $f_d$ is chosen. 

\begin{figure}[t!]
  \centering
  \includegraphics[width=\columnwidth]{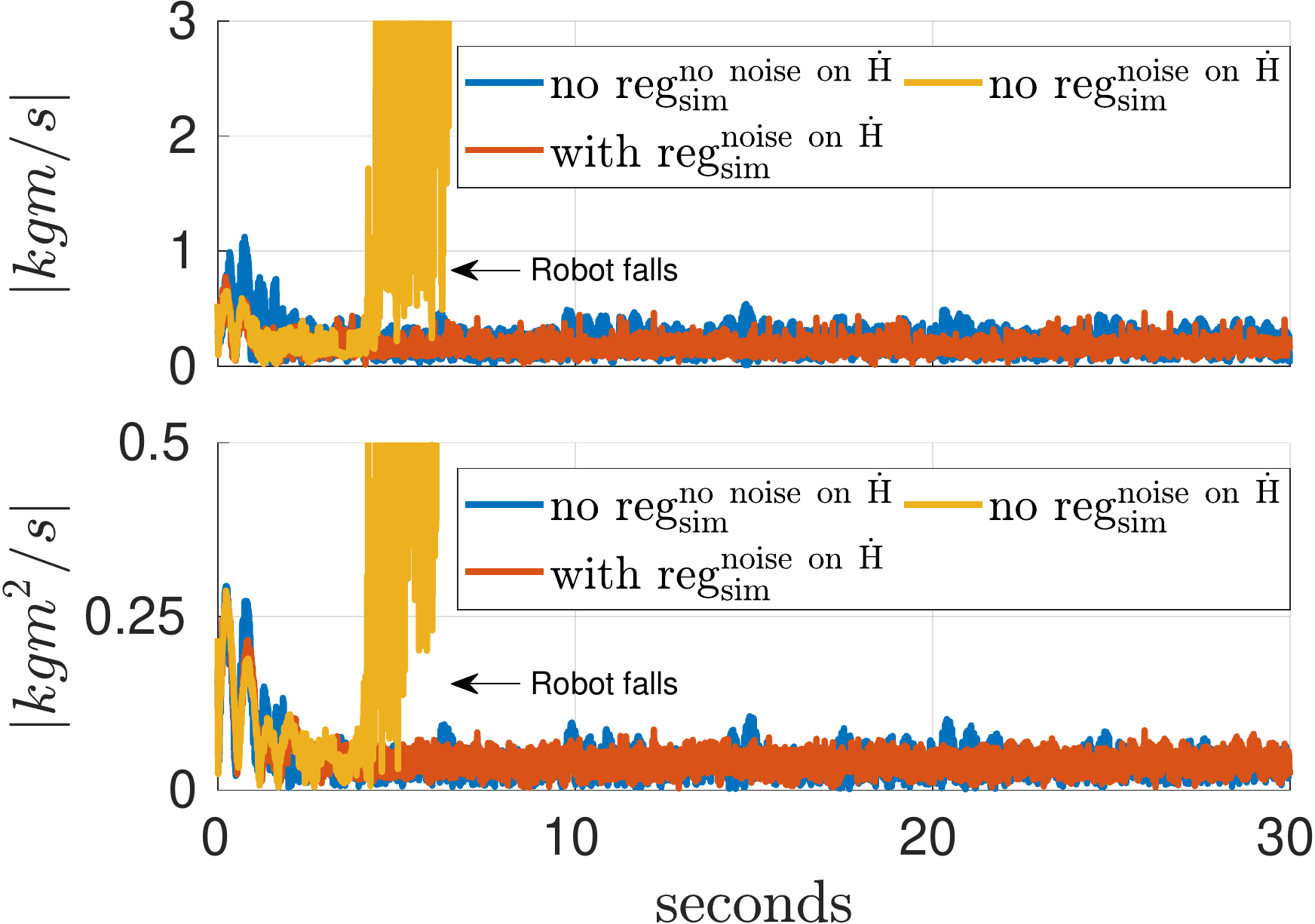}
  \caption{Linear (top) and angular (bottom) momentum error norm during two feet balancing simulations; dynamic model overestimated by $7.5 \%$. 
  No regularization are added when the momentum rate of change is not affected by errors. If $\dot{H}$ is biased, the robot falls unless the  regularization is used.}
  \label{fig:mom_err_norm_reg_dynamic}
\end{figure}

It is important to point out that the regularization term $-k_e(\xi -\xi_d)$ is only necessary in case of errors when estimating the momentum rate-of-change $\dot{H}$. To support this statement, 
%
Figure \ref{fig:mom_err_norm_reg_dynamic} shows the linear and angular momentum error norm during balancing simulations when $7.5 \%$ modeling errors are added to 
the parameters $M,m,C$ and $G$. 
Note that the robot falls after few seconds if no regularization is added to Eq.~\eqref{eq:input_momentum} (orange line). When the regularization term is added (red line) or the errors on $\dot{H}$ are removed (blue line), stability is retained. In this stability retained case, let us remark that only the estimation of $\dot{H}$ is evaluated correctly, while the model errors are kept present for all the other calculations. 

We also carried out robustness tests on the real iCub during a contact switching scenario \cite{nava2016}. The robot starts balancing on two feet, then it switches to balance on the left foot via a finite state machine, and performs highly dynamic movements on the left foot. Finally, the robot returns back to two feet balancing. Results are reported in Table \ref{tab:robustness}: the robot succeeded to conclude the demo $60\%$ of times in case of parameters overestimation by $7.5 \%$. Dealing with parameters underestimation, instead, seems to be a more challenging task for the controller~\eqref{eq:input_momentum_modif} 
despite the presence of the regularization term.

%
\begin{figure}[t]
  \centering
  \includegraphics[width=\columnwidth]{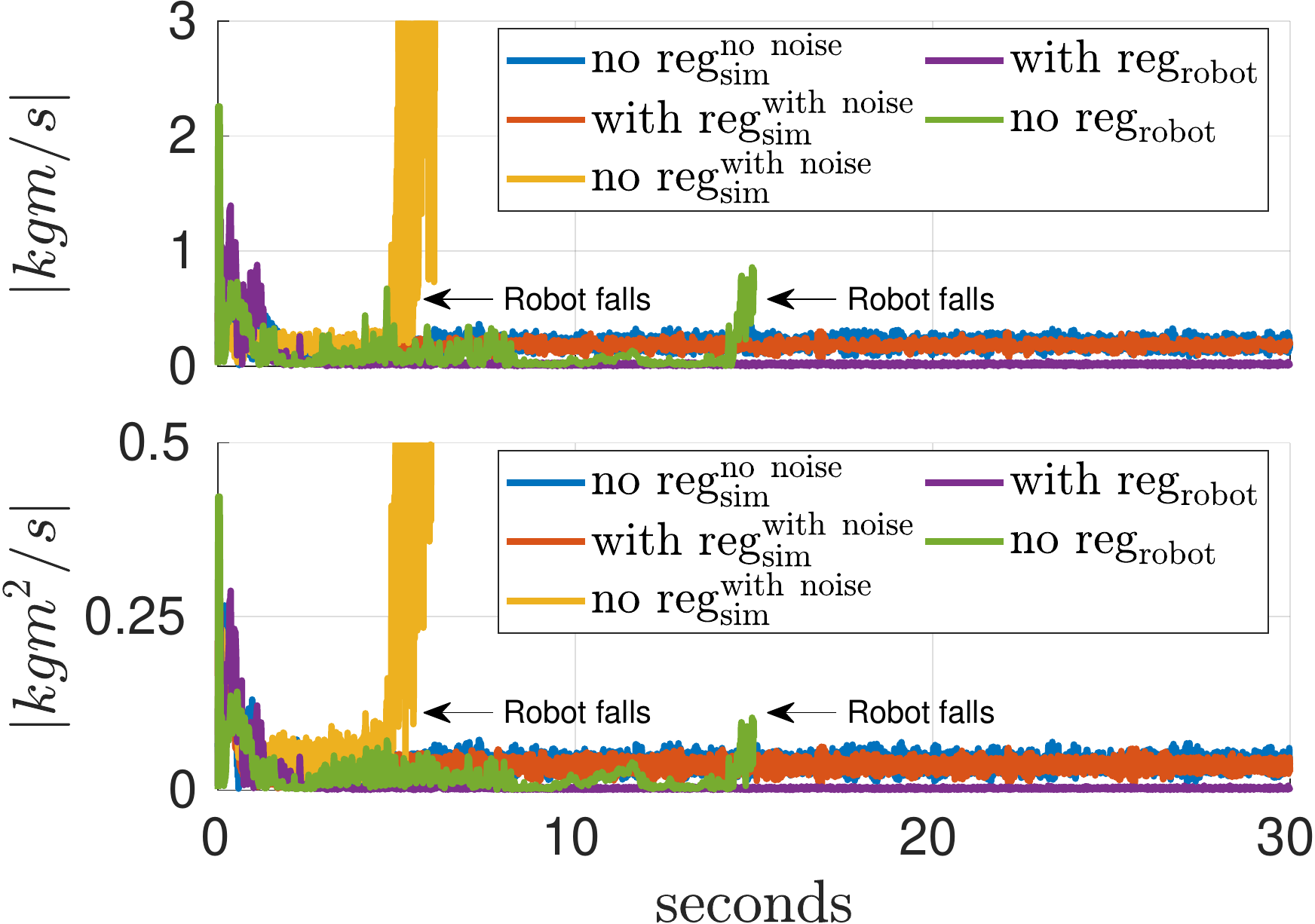}
  \caption{Linear (top) and angular (bottom) momentum error norm during two feet balancing. On the real robot the additional regularization term is required, while in simulation it is not. Adding noise to the FT  measurements in simulation generates a response similar to that of the real iCub.}
  \label{fig:mom_err_norm_reg}
\end{figure}

When mounted on the robot, the FT sensors accuracy is affected by several phenomena such as temperature, internal stresses and vibrations. More specifically, we observed that even after FT sensor fine calibration the linear forces measurements still have an offset of $\pm 2.5 N$, 
and this offset biases the estimation of $\dot{H}$.
\begin{table}[t!]
\centering
\caption{Robustness tests on the real iCub while performing highly dynamic movements.}
\begin{tabular}{|c|c|}
\hline
\multicolumn{2}{|c|}{\textbf{Robustness w.r.t. modeling errors on real robot}} \\ \hline
Error on dynamic model $[\%]$ & Success rate $[\rm trials]$ \\ \hline
Overest.  $5\%$   & $5/5$         \\ \hline
Overest.  $7.5\%$ & $3/5$         \\ \hline
Overest.  $10\%$  & $1/5$         \\ \hline
Underest. $5\%$   & $1/5$         \\ \hline
\end{tabular}
\label{tab:robustness}
\vspace{-0.5cm}
\end{table}
Figure \ref{fig:mom_err_norm_reg} shows the behavior of the linear and angular momentum error norm during several two feet balancing simulations and experiments. On the real iCub, the robot falls after few seconds when $k_e = 0$, as pointed out by the \emph{green} line. When adding the regularization term in Eq. \eqref{eq:input_momentum_modif}, stability is retained and the momentum error does not diverge (\emph{purple} line). On the other hand, the \emph{blue} line is obtained in simulation with perfect estimation of  external forces, and with $k_e = 0$. In this case, the momentum error does not diverge, thus showing again that the regularization term is not needed when $\dot{H}$ is properly estimated. In simulation, we injected a constant offset of amplitude $2.5 N$ to the "measured"  $f_x$ component of one of the two contact wrenches. Results correspond to the \emph{orange} line in Figure \ref{fig:mom_err_norm_reg}: stability is no longer retained and the robot falls after 
few seconds. With the regularization term, the previous balancing performances are restored (\emph{red} line).

\subsection{Comparison with a momentum-based QP controller}


We compared the performance of the \emph{momentum-based} jerk controller~\eqref{eq:input_momentum_modif} with a classical \emph{momentum-based} QP controller that solves the optimization problem \eqref{eq:sot} on the real iCub during the contact switching scenario 
introduced in Section \ref{subsec:regularization}.
%
Both controllers have been fine tuned for the specific demo. The goal is to show that the momentum-based jerk control  guarantees performances that are comparable with a controller already available in the literature. Also, the momentum-based jerk control provides smoother references to the torque controller, as the desired contact wrenches $f^*$ are always continuous.
%

\begin{figure}[t!]
  \centering
  \includegraphics[width=\columnwidth]{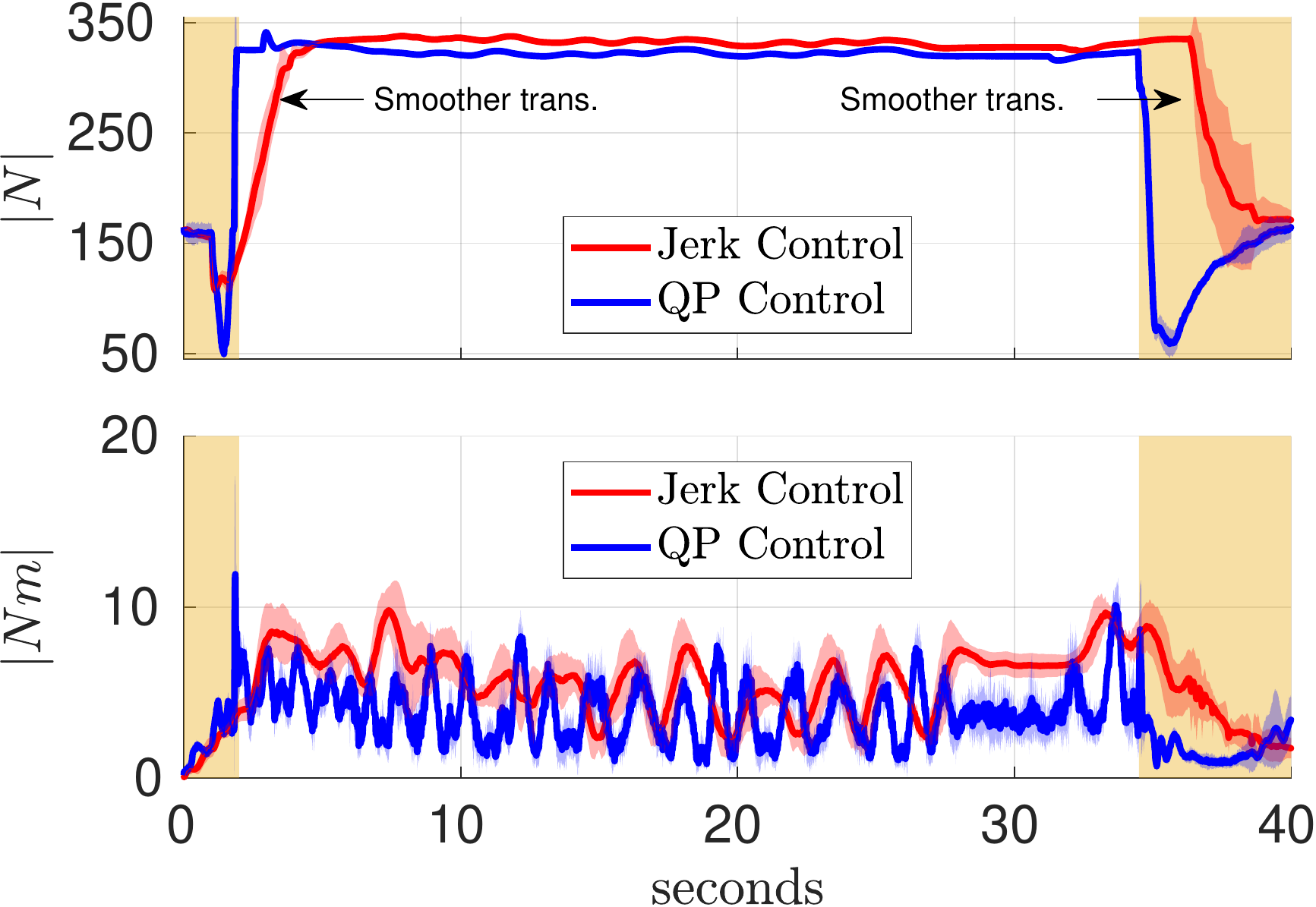}
  \vspace*{-0.27cm}
  \caption{Norm of the left foot input forces (top) and moments (bottom). The jerk control (red line) helps in providing smoother references to the torque controller during contact switching (orange background).} 
  \label{fig:des_force_torque_norm}
  \vspace{0.25cm}
\end{figure}

Figure \ref{fig:des_force_torque_norm} depicts the norm of the left foot (input) contact forces and moments for both the momentum-based jerk control and the momentum-based QP control. Results have been achieved by running 10 experiments for each control strategy. The solid lines represents the average values, while the transparent regions are the variance  over the 10 experiments. The orange background represents the instants at which the robot is balancing on two feet, while the white background is when the robot is balancing on the left foot. 
%
%

The momentum-based jerk controller helps provide smoother references to the torque controller during transitions. In Figure \ref{fig:mom_err_norm}, we compared the norms of the linear and angular momentum error. During transition from two feet to left foot balancing, a peak of error is present when the momentum-QP control is used. The peak is caused by the sharp change in the input forces. When jerk control is used (and smoother force input is required), the peak error is reduced by ~$90\%$. During highly dynamic movements (white background), jerk and momentum-based QP control show similar tracking performances.
%

%
Both controllers use Eq. \eqref{eq:input_torques} to generate the input torques. Figure \ref{fig:joint_err_norm} verifies the boundedness of the system's \emph{zero dynamics}. In both cases, the zero dynamics does not diverge. Convergence to zero of the joint position error is not necessarily expected as the controllers implement strict task priorities, and the \emph{postural task} is chosen as the lowest priority task. For further details, a video of the experiment is attached to this paper. 

\subsection{Disturbance rejection}

To evaluate the robustness of the momentum-based jerk controller against unmodeled external force perturbations, we performed the following experiment. The robot balances on two feet and a person pushes and pulls continuously the robot's upper body. The applied external force is unmodeled, so it is treated as a disturbance by the momentum-based jerk controller. Figure \ref{fig:mom_err_norm_disturbances} shows the momentum rate of change error norm and the momentum error norm during interaction. Despite the high peaks of errors when the external force is applied, the controller is still able to retain stability. When the force is removed, in fact, the momentum error and its rate of change converge to a value. Exact convergence to zero of the momentum derivative error on the real iCub is difficult to obtain because of the low sensitivity of the FT sensors, so a compromise is achieved by properly tuning the corresponding feedback gains. A video describing the experiment is attached to the paper.

%% file: tex/conclusions.tex
\section{Conclusions and Future Work}
\label{sec:conclusion}

In this paper, we addressed some common limitations of force and torque controllers for floating base systems based on Quadratic Programming. More specifically, we removed inequality constraints from the optimization problem by designing an invertible, one-to-one mapping that parametrises the contact wrenches into a new set of unconstrained variables. This parametrization guarantees that the contact wrenches always satisfy the contact stability constraints. Based on this mapping, we designed a  jerk control framework for floating base systems. We then analyzed a specific use case of the jerk controller, namely a momentum-based jerk control architecture for balancing a 23 DoFs humanoid robot. The controller has been validated both in simulation and on the real iCub, and compared with a classical momentum-based controller. 
%
%
Sensitivity to errors in the momentum rate of change estimation is identified as a drawback of the  approach, as it may affect stability and convergence of the closed loop dynamics. A solution for increasing robustness of the controller w.r.t. biases on momentum estimation is presented.

A limitation is that the proposed jerk control architecture does not take into account joint position and torque limits. A future work may involve the integration of these limits in the control framework by extending the approach presented in \cite{marie2016} to the case of floating base robots.
%
\begin{figure}[t!]
  \centering
  \includegraphics[width=\columnwidth]{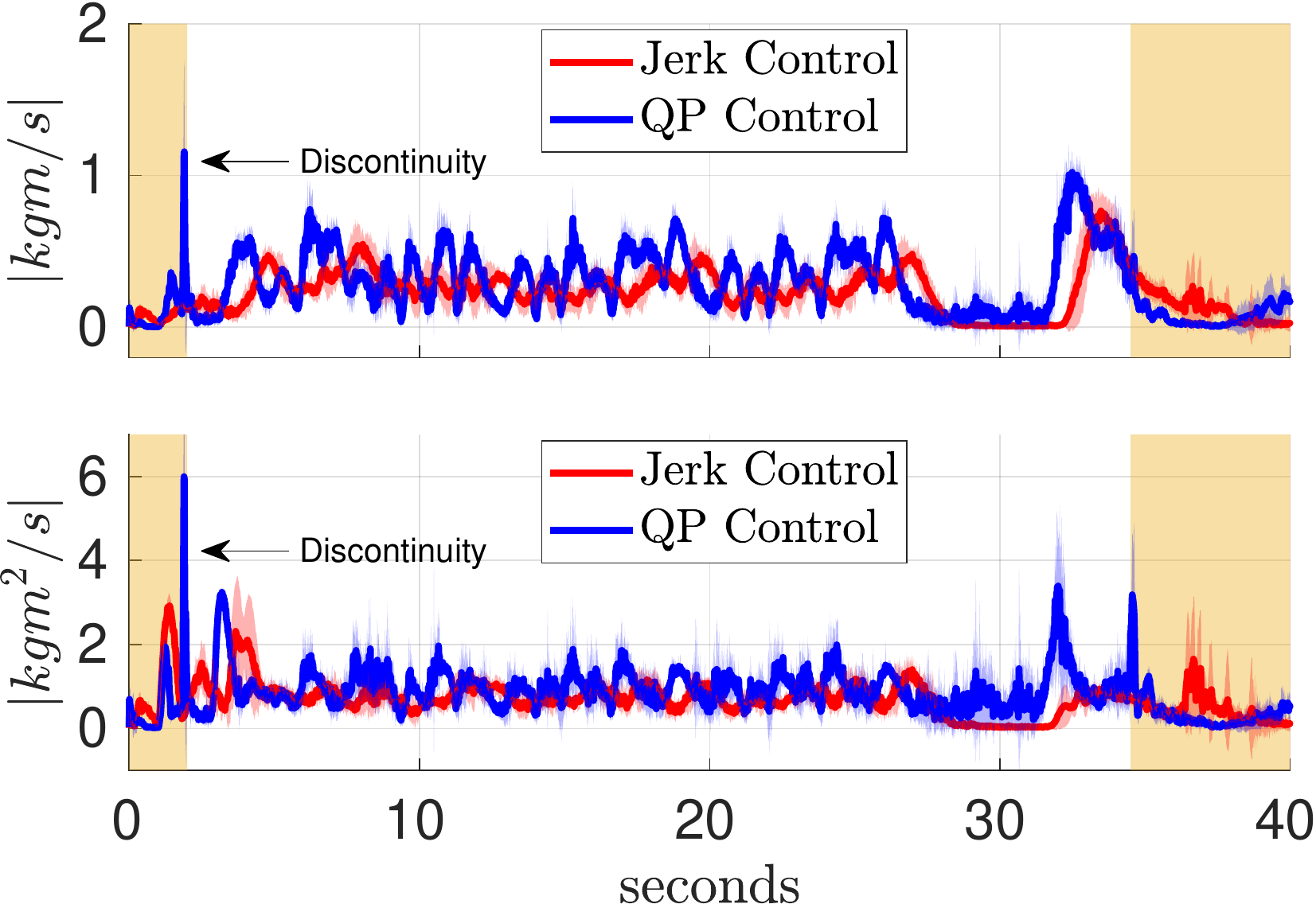}
  \caption{Linear (top) and angular (bottom) momentum error norm during contact switching demo. A peak of momentum error is present during contact switching if momentum-QP control is used. During highly dynamic movements, the tracking performances of the two controllers are comparable.}
  \label{fig:mom_err_norm}
\end{figure} 
%
In this paper, the jerk control architecture implements a momentum task and a postural task for balancing. Further development will be done to extend the jerk control to more complex control objectives, as for example humanoid robot walking.

%% file: tex/appendix.tex
\appendix
  
  \subsection{Proof of Lemma \ref{lemma_1}}
  \label{appendix_1}
\noindent
  \textbf{Proof of 1):} when $f^k =  \phi(\xi^k)$, the constraint 
  on 
  the vertical force Eq. \eqref{eq:pos_vert_forces} is given by:
  \begin{equation*}
        f_z = e^{\xi_3} + f_z^{min} > f_z^{min},
  \end{equation*}
  which is satisfied for all $\xi_3$.
  Now, substitute the  parametrization~\eqref{eq:parametrized_wrench} in the remaining stability constraints \eqref{eq:static_friction}--\eqref{eq:torsional_friction}:
  \begin{IEEEeqnarray}{RCL}
  \IEEEyesnumber
    \label{eq:contact_constraints_parametrized}
    \sqrt[]{{\mu_c}^2\frac{\tanh^2(\xi_1) \,f_z^2}{{1+\tanh^2(\xi_2)}} +   {\mu_c}^2\frac{\tanh^2(\xi_2) \,f_z^2}{{1+\tanh^2(\xi_1)}}} &<& {\mu_c} f_z \quad \quad \IEEEyessubnumber   \label{eq:static_friction_parametrized},\\ 
     y_c^{min} < \frac{(\delta_y \tanh(\xi_4) + \delta_{y_0})\, f_z}{f_z} < y_c^{max} \IEEEyessubnumber \label{eq:CoP_y_parametrized},\\ 
     -x_c^{max} < \frac{(\delta_x \tanh(\xi_5) + \delta_{x_0})\, f_z}{f_z} < -x_c^{min} \IEEEyessubnumber \label{eq:CoP_x_parametrized},\\ 
    \left|\frac{\mu_z \tanh(\xi_6)\, f_z}{f_z}\right| \hspace{0.5 mm} < \mu_z. \IEEEyessubnumber \label{eq:torsional_friction_parametrized}
  \end{IEEEeqnarray}
  \begin{figure}[t]
   \centering
   \includegraphics[width=\columnwidth]{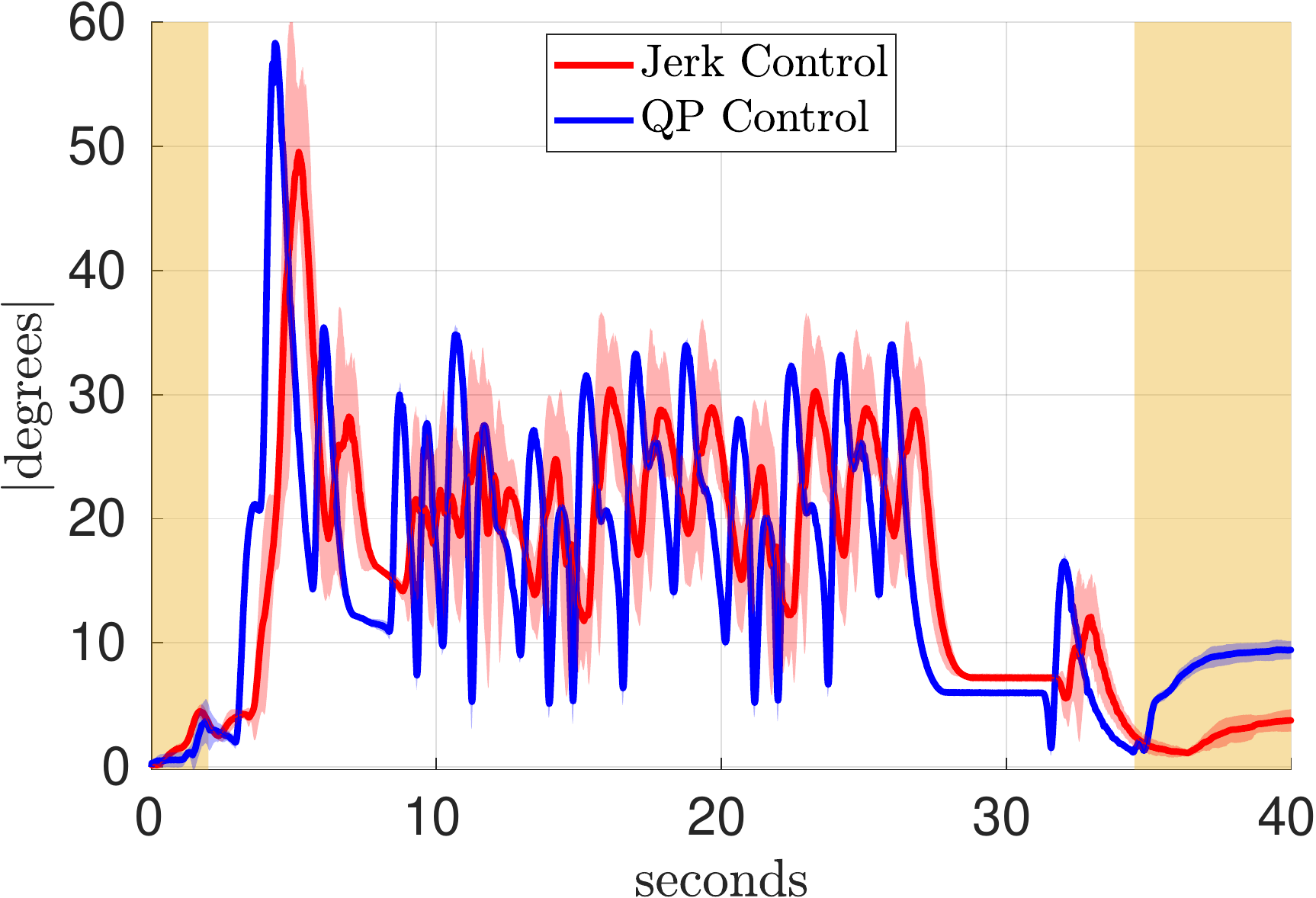}
   \caption{Joint position error norm during highly dynamics movements. The plot shows that the system's zero dynamics does not diverge while achieving the primary task.}
   \label{fig:joint_err_norm}
   \vspace{-0.5cm}
  \end{figure} 
    \begin{figure}[t!]
    \centering
    \includegraphics[width=\columnwidth]{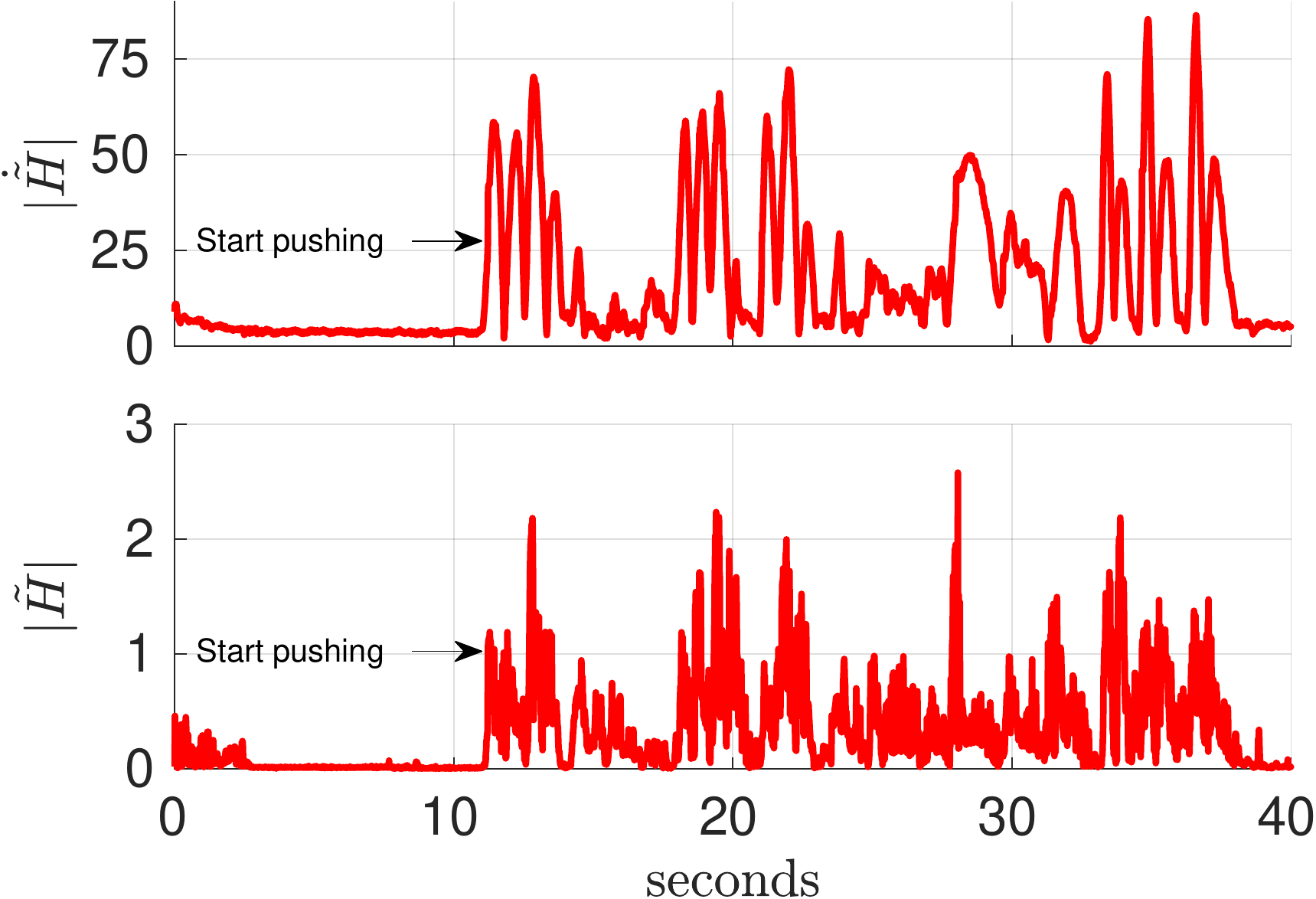}
     \caption{Momentum rate of change (top plot) and momentum error norm (Bottom plot) during the disturbance rejection experiment. The controller can retain stability despite the action of the external unmodeled forces.}
     \label{fig:mom_err_norm_disturbances}  
  \end{figure}
  
  The vertical force $f_z$ is greater than zero and can be removed from  \eqref{eq:contact_constraints_parametrized}, thus leading to the following set of  inequalities:
  \begin{IEEEeqnarray}{LCL}
  \IEEEyesnumber
    \label{eq:contact_constraints_parametrized_simplified}
    \sqrt{\frac{\tanh^2(\xi_1)}{1+\tanh^2(\xi_2)} + \frac{\tanh^2(\xi_2)}{1+\tanh^2(\xi_1)}} < 1 \IEEEyessubnumber   \label{eq:static_friction_parametrized_simplified},\\ 
     y_c^{min} < \delta_y \tanh(\xi_4) + \delta_{y_0} < y_c^{max} \IEEEyessubnumber \label{eq:CoP_y_parametrized_simplified},\\ 
     -x_c^{max} < \delta_x \tanh(\xi_5) + \delta_{x_0} < -x_c^{min} \IEEEyessubnumber \label{eq:CoP_x_parametrized_simplified},\\ 
    |\tanh(\xi_6)| \hspace{0.5 mm}  < 1, \IEEEyessubnumber \label{eq:torsional_friction_parametrized_simplified}
  \end{IEEEeqnarray}
  where also the coefficients $\mu_c$ and $\mu_z$ have been removed from Eq. \eqref{eq:static_friction_parametrized} and \eqref{eq:torsional_friction_parametrized}, respectively. It is straightforward to verify that the constraint \eqref{eq:torsional_friction_parametrized_simplified} is verified for all finite $\xi_6$.
  Also, direct calculations on Eq. \eqref{eq:CoP_y_parametrized_simplified}-\eqref{eq:CoP_x_parametrized_simplified} show that:
  \begin{IEEEeqnarray}{LCL}
    \label{eq:cop_constraints_simplified}
     y_c^{min} < \frac{y_c^{max}(1+ \tanh(\xi_4)) + y_c^{min}(1- \tanh(\xi_4))}{2} < y_c^{max} \nonumber \\ 
     x_c^{min} < \frac{x_c^{max}(1- \tanh(\xi_5)) + x_c^{min}(1+ \tanh(\xi_5))}{2} < x_c^{max}. \nonumber
  \end{IEEEeqnarray}
  Using the facts that $y_c^{min} < y_c^{max}$, $x_c^{min} <  x_c^{max}$ and $|\tanh(\cdot)| < 1$,  one easily shows that the above quantities are always satisfied for all $\xi_4,\xi_5$. Concerning the remaining constraint Eq. \eqref{eq:static_friction_parametrized_simplified}, note that the argument of the square root can be rearranged as follows:
  \begin{IEEEeqnarray}{LCL}
    \frac{\tanh^2(\xi_1) (1 + \tanh^2(\xi_1)) + \tanh^2(\xi_2) (1 + \tanh^2(\xi_2))}{\tanh^2(\xi_1)+ \tanh^2(\xi_2) + \tanh^2(\xi_1)\tanh^2(\xi_2) +1}  \nonumber \\ 
    = \frac{\tanh^2(\xi_1) + \tanh^2(\xi_2) + \tanh^4(\xi_1) + \tanh^4(\xi_2)}{\tanh^2(\xi_1)+ \tanh^2(\xi_2) + \tanh^2(\xi_1)\tanh^2(\xi_2) +1}. \nonumber
  \end{IEEEeqnarray}
  So, the constraint \eqref{eq:static_friction_parametrized_simplified} is satisfied if and only if
    \begin{IEEEeqnarray}{LCL}
    \label{constraint28a-2}
    \tanh^2(\xi_1)\tanh^2(\xi_2) - \tanh^4(\xi_1) - \tanh^4(\xi_2) + 1 > 0.  \IEEEeqnarraynumspace
  \end{IEEEeqnarray}
  Since $|\tanh(\cdot)| < 1$, then
    $\tanh^4(\cdot) \leq \tanh^2(\cdot)$. 
    As a consequence, one easily finds a minorant of the left hand side of~\eqref{constraint28a-2}, namely 
   \begin{IEEEeqnarray}{LCL}
    \label{constraint28a-3}
    \tanh^2(\xi_1)\tanh^2(\xi_2) - \tanh^4(\xi_1) - \tanh^4(\xi_2) + 1 &\geq& \nonumber \\ \tanh^2(\xi_1)\tanh^2(\xi_2) - \tanh^2(\xi_1) - \tanh^2(\xi_2) + 1 &=& \nonumber \\ 
    \left(1-\tanh^2(\xi_1)\right)\left( 1-\tanh^2(\xi_2)\right).
    \IEEEeqnarraynumspace 
  \end{IEEEeqnarray}
Since $|\tanh(\cdot)| < 1$, then~\eqref{constraint28a-3} is (strictly) greater than zero, which renders~\eqref{constraint28a-2} satisfied.

\noindent
  \textbf{Proof of 2):} First, it follows from \eqref{eq:parametrized_wrench} that  any value of $\xi$ generates a unique $f^k \in \mathcal{K'}$. Now, assume that $f^k \in  \mathcal{K'}$. Then, we have to show that there exists a unique $\xi = \phi^{-1}(f^k)$. It is straightforward to compute the \emph{inverse mapping} of the vertical force and moments parametrization:
  \begin{IEEEeqnarray}{LCL}
        \xi_3 = \ln{}\bigg(f_z -f_z^{min}\bigg) \nonumber\\
        \xi_4 = \text{atanh}\bigg(\frac{M_x - \delta_{y_0}f_z}{\delta_{y}f_z}\bigg) \nonumber \\
        \xi_5 = \text{atanh}\bigg(\frac{M_y - \delta_{x_0}f_z}{\delta_{x}f_z}\bigg) \nonumber \\
        \xi_6 = \text{atanh}\bigg(\frac{M_z}{\mu_{z}f_z}\bigg). \nonumber
  \end{IEEEeqnarray}
 Since $\mathcal{K'} \subset \mathcal{K}$, then $f^k$ satisfies \eqref{eq:contact_constraints}. Using this fact and the definitions \eqref{eq:feet_size_constraints} we conclude that the solutions $(\xi_3,\xi_4,\xi_5,\xi_6)$ exist. Furthermore, since the above equations are composed of one-to-one correspondences (hyperbolic tangent, logarithm), then these solutions are also unique.

  For what concerns the tangential forces $f_x$ and $f_y$, let us recall the expressions of the $f_x$ and $f_y$ parametrizations:
  \begin{IEEEeqnarray}{LCL}
  \IEEEyesnumber
  \label{eq:parametrization_f_xy}
   f_x = {\mu_c}\frac{\tanh(\xi_1) \,f_z}{\sqrt{{1+\tanh^2(\xi_2)}}} \IEEEyessubnumber\\
   f_y = {\mu_c}\frac{\tanh(\xi_2) \,f_z}{\sqrt{{1+\tanh^2(\xi_1)}}}. \IEEEyessubnumber
  \end{IEEEeqnarray}
  An easy way to compute the inverse mapping is to raise to the square Eq. \eqref{eq:parametrization_f_xy}, which gives:
  \begin{IEEEeqnarray}{LCL}
  \IEEEyesnumber
  \label{eq:parametrization_f_xy_sq}
   f_x^2 = {\mu_c}^2\frac{\tanh^2(\xi_1) \,f_z^2}{{1+\tanh^2(\xi_2)}} \IEEEyessubnumber \\
   f_y^2 = {\mu_c}^2\frac{\tanh^2(\xi_2) \,f_z^2}{{1+\tanh^2(\xi_1)}}. \IEEEyessubnumber
  \end{IEEEeqnarray}
  In the resulting equations, the square hyperbolic tangents $\tanh^2(\xi_1)$, $\tanh^2(\xi_2)$ appear linearly. Therefore they can be easily computed through matrix inversion:
  \begin{IEEEeqnarray}{LCL}
  \IEEEyesnumber
   \label{eq:inverse_mapping_xy}
   \begin{bmatrix}
     \tanh^2(\xi_1) \\
     \tanh^2(\xi_2)
   \end{bmatrix} &=& \begin{bmatrix}
                        {\mu_c}^2f_z^2 & -f_x^2 \\
                        -f_y^2 & {\mu_c}^2f_z^2 
                      \end{bmatrix}^{-1} \begin{bmatrix}
                                            f_x^2 \\
                                            f_y^2
                                          \end{bmatrix}.
  \end{IEEEeqnarray}
   Since $f^k \in \mathcal{K'}$, then the right hand side of \eqref{eq:inverse_mapping_xy} is necessarily smaller than one and there exists at least one $\xi$ satisfying \eqref{eq:inverse_mapping_xy}. Resolving Eq. \eqref{eq:inverse_mapping_xy} w.r.t. $\xi_1,\xi_2$ gives \emph{two} possible solutions, namely $\xi_{1(2)} = \pm \text{atanh}\left(\sqrt{\tanh^2(\xi_{1(2)})}\right)$. However, only one of the two solutions satisfies the parametrization \eqref{eq:parametrization_f_xy}: in fact, the terms $f_z, \mu_c$, and $\sqrt{{1+\tanh^2(\xi_{1(2)})}}$ on the right-hand side of \eqref{eq:parametrization_f_xy} are always positive. So, the sign of $\xi_1$ ($\xi_2$) must correspond to the sign of $f_x$ ($f_y$), leading to the unique solution:
   \begin{IEEEeqnarray}{LCL}
    \xi_{1} = \text{sign}(f_x) \text{atanh}\left(\sqrt{\tanh^2(\xi_{1})}\right) \nonumber \\
    \xi_{2} = \text{sign}(f_y) \text{atanh}\left(\sqrt{\tanh^2(\xi_{2})}\right). \nonumber
  \end{IEEEeqnarray}
  
  \textit{Remark:} it is possible to verify that if $f_x$ and $f_y$ belong to $\mathcal{K'}$, then the matrix inversion in Eq. \eqref{eq:inverse_mapping_xy} can always be performed. In fact, singularities arise when \[\det\left(\begin{bmatrix}
                        {\mu_c}^2f_z^2 & -f_x^2 \\
                        -f_y^2 & {\mu_c}^2f_z^2 
                      \end{bmatrix}\right) = \mu_c^4f_z^4 -f_x^2f_y^2 = 0.\] Substituting Eq. \eqref{eq:parametrization_f_xy_sq} in the expression of the determinant allows to verify that the condition $\mu_c^4f_z^4 = f_x^2f_y^2$ never occurs for any $\xi_1,\xi_2$.
  
  \noindent
  \textbf{Proof of 3):} let $\Phi_k \in \mathbb{R}^{6 \times 6}$ denote the gradient of $f^k =  \phi(\xi)$. Then, straightforward calculations show that $\Phi_k$ is given by:
  \begin{equation}
     \label{gradient_matrix}
     \Phi_k = 
     \begin{bmatrix}
     F_{11} & F_{12} & F_{13}  & 0 & 0 & 0      \\
     F_{21} & F_{22} & F_{23}  & 0 & 0 & 0      \\
     0      & 0      & F_{33}  & 0 & 0 & 0      \\
     0      & 0      & F_{43}  & F_{44} & 0 & 0 \\
     0      & 0      & F_{53}  & 0 & F_{55} & 0 \\
     0      & 0      & F_{63}  & 0 & 0 & F_{66} \\
     \end{bmatrix}. \nonumber
  \end{equation}
  Applying the Laplace's formula for the calculation of the determinant of $\Phi_k$ leads to:
  \begin{equation}
     \text{det}(\Phi_k) = F_{66}F_{55}F_{44}F_{33} \text{det}\begin{bmatrix} F_{11} & F_{12} \\
     F_{21} & F_{22} \end{bmatrix}  \nonumber
  \end{equation}
  where one has:
  \begin{align}
    F_{11} =\,\, & \frac{{\mu_c}(1-\tanh^2(\xi_{1}))(e^{\xi_3} + f_z^{min})}{\sqrt{1 +\tanh^{2}(\xi_{2})}}, \nonumber \\ 
    F_{12} =\,\, & \frac{{\mu_c}\tanh(\xi_{1})(e^{\xi_3} + f_z^{min})}{(1 +\tanh^{2}(\xi_{2}))^{\frac{3}{2}}}(\tanh^3(\xi_{2})-\tanh(\xi_{2})) , \nonumber \\
    F_{21} =\,\, & \frac{{\mu_c}\tanh(\xi_{2})(e^{\xi_3} + f_z^{min})}{(1 +\tanh^{2}(\xi_{1}))^{\frac{3}{2}}}(\tanh^3(\xi_{1})-\tanh(\xi_{1})) ,  \nonumber \\
    F_{22} =\,\, & \frac{{\mu_c}(1-\tanh^2(\xi_{2}))(e^{\xi_3} + f_z^{min})}{\sqrt{1 +\tanh^{2}(\xi_{1})}}, \nonumber \\
    F_{33} = \,\,& e^{\xi_{3}}, \nonumber \\
    F_{44} = \,\,& \delta_y (1 - \tanh^2(\xi_4))(e^{\xi_3} + f_z^{min}), \nonumber \\
    F_{55} = \,\,& \delta_x (1 - \tanh^2(\xi_5))(e^{\xi_3} + f_z^{min}), \nonumber \\
    F_{66} = \,\,& \mu_{z}(1 - \tanh^2(\xi_6))(e^{\xi_3} + f_z^{min}).  \nonumber
  \end{align}
  It can be verified that $F_{33}, F_{44}, F_{55}$ and $F_{66}$ are always different from zero for any finite $\xi_3,\xi_4,\xi_5,\xi_6$. We are then left to evaluate the determinant of $\begin{bmatrix} F_{11} & F_{12} \\ F_{21} & F_{22} \end{bmatrix}$, which is $F_{11}F_{22} - F_{12}F_{21}$. After noting that the product $F_{11}F_{22}$ is contained in the expression of $F_{12}F_{21}$, one is left with:
  \begin{IEEEeqnarray}{LCL}
    \text{det} \begin{bmatrix} F_{11} & F_{12} \\
                               F_{21} & F_{22} \end{bmatrix} \nonumber &=&
      {\mu_c^2}(e^{\xi_3} + f_z^{min})^2  \cdot \nonumber \\ 
      & & \frac{(1-\tanh^2(\xi_{1}))(1-\tanh^2(\xi_{2}))}{\sqrt{(1+\tanh^2(\xi_{1}))(1+\tanh^2(\xi_{2}))}} \cdot \nonumber \\
      & &  \frac{(1 + \tanh^2(\xi_{1}) + \tanh^2(\xi_{2}))}{(1+\tanh^2(\xi_{1}))(1+\tanh^2(\xi_{2}))} \nonumber
  \end{IEEEeqnarray}
  which is non-zero for any finite $\xi_1, \xi_2, \xi_3$. Hence, matrix $\Phi_k$ is invertible for any finite $\xi$.

\subsection{Proof of Lemma \ref{lemma_12}}
\label{appendix_12}

\noindent
 \textbf{Proof of 1):} first, observe that Lyapunov stability of the equilibrium point  $(\tilde{H},\dot{\tilde{H}}) = (0,0)$ follows from \eqref{eq:closed_loop_fb_linH} with $\ddot{H}^*$ given by \eqref{eq:closed_loop_fb_lin}. By definition, Lyapunov stability implies that there exist initial conditions $(\tilde{H},\dot{\tilde{H}})(0)$ such that the system trajectories $(\tilde{H},\dot{\tilde{H}})(t)$ remain as close as we wish to the point $(0,0)$. We are then left to show that the variable $\xi$ remains bounded while the system trajectories evolve close to the equilibrium point: the boundedness of $\xi$ would imply that the matrix $A\Phi(\xi)$ in \eqref{eq:centroidal_momentum_acc} remains of full rank, thus ensuring that the equality \eqref{eq:closed_loop_fb_linH} can be satisfied $\forall t$. 
 
Now, since the desired momentum $H_d$ is a feasible system equilibrium, then there exists a bounded $\xi_e(t)$ such that 
  \begin{IEEEeqnarray}{LCL}
  \label{equilibrium}
    \dot{H}_d &=& A\phi(\xi_e) -mge_3.
  \end{IEEEeqnarray}
Furthermore, the behaviour of system 
\[\dot{\tilde{H}} = A\phi(\xi) -mge_3 -\dot{{H}}_d\] 
close to the equlibrium point can be obtained via linearisation techniques, which in view of \eqref{equilibrium} yields
  \begin{IEEEeqnarray}{LCL}
  \label{linearisation}
    \dot{\tilde{H}} = A(\phi(\xi_e) + \Phi(\xi_e)\tilde{\xi}) -mge_3 - \dot{{H}}_d = A\Phi(\xi_e)\tilde{\xi}, \IEEEeqnarraynumspace
  \end{IEEEeqnarray}
with $\tilde{\xi} = \xi - \xi_e$. 
Since it is assumed that the robot makes a single contact with the environment, namely $n_c = 1$, then the matrix $A \in \mathbb{R}^{6\times6}$. From \eqref{matrix_Ak} one easily verifies that the matrix $A$ is always invertible, with $A^{-1}$ a bounded matrix. Then, one has \[|\tilde{\xi}| = |(A\Phi(\xi_e))^{-1}\dot{\tilde{H}}| \leq c_a|\dot{\tilde{H}}|,\]
with $c_a > 0$. As a consequence of the Lyapunov stability of $(\tilde{H},\dot{\tilde{H}}) = (0,0)$,  there exist initial conditions $(\tilde{H},\dot{\tilde{H}})(0)$ for which the variable $\tilde{\xi}$ stays bounded and as close as we wish to zero, thus rendering the linearisation \eqref{linearisation} consistent $\forall \ t$. Finally, since $\xi_e$ is bounded, then $\xi$ is also bounded, and the equality \eqref{eq:closed_loop_fb_linH} can be satisfied $\forall t$. This latter fact, along with $\ddot{H}^*$ given by \eqref{eq:closed_loop_fb_lin}, then implies local asymptotic stability since \eqref{eq:closed_loop_fb_linH} is satisfied $\forall \ t$.

\noindent
\textbf{Proof of 2):} since it is assumed that the variable $\xi$ is bounded $\forall t$, then \eqref{eq:closed_loop_fb_linH} can be satisfied $\forall \ t$ because the matrix $A\Phi(\xi)$ in \eqref{eq:centroidal_momentum_acc} remains of full rank. Then, the closed loop dynamics is $\ddot{\tilde{H}} = - K_d\dot{\tilde{H}} - K_p{\tilde{H}} \ \forall t$, which implies global asymptotic stability of the equilibrium point  $(\tilde{H},\dot{\tilde{H}}) = (0,0)$. 

  \subsection{Proof of Lemma \ref{lemma_2}}
  \label{appendix_2}

\noindent
  \textbf{Proof of 1):} define the following Lyapunov function candidate:
   \begin{equation}
   \label{lyapunov-integral}
      V(I, \tilde{H}, \zeta):= \frac{1}{2}I^{\top}K_pI + \frac{1}{2} \tilde{H}^\top\tilde{H} + \frac{1}{2}\zeta^\top K_0 \zeta. 
   \end{equation}
   Note that $V = 0 \iff (I, \tilde{H}, \zeta) = (0,0,0)$. Compute the Lyapunov function derivative $\dot{V}$:
   \begin{align}
     \dot{V} & = I^\top K_p \tilde{H} + \tilde{H}^\top \dot{\tilde{H}} + \zeta^\top K_o \dot{\zeta} \nonumber \\
     & = I^\top K_p \tilde{H} + \tilde{H}^\top (\zeta -K_p I -K_d\tilde{H}) + \zeta^\top K_o \dot{\zeta} \nonumber \\
     & = - \tilde{H}^\top K_d\tilde{H} + \zeta^\top K_o(\dot{\zeta} + K_o^{-1}\tilde{H}). \nonumber
   \end{align}
   It is easy to prove that $\dot{V} \leq 0$ when $\dot{\zeta} + K_o^{-1}\tilde{H} = -\zeta$. 
   Then, in view of 
   $\dot{\zeta}$ 
   given by \eqref{eq:dot_zeta} and the definition of $\zeta$ one has:
   \begin{align}
   \label{eq:input_momentum_definition}
    \dot{A}f + A\Phi\dot{\xi} - \ddot{H}_d + K_d \dot{\tilde{H}} + K_p\tilde{H} + K_o^{-1}\tilde{H} \\
    = -Af +mge_3 + \dot{H}_d - K_d \tilde{H} - K_p I, \nonumber
   \end{align}  
   and after a rearrangement, Eq. \eqref{eq:input_momentum_definition} leads to the definition of the control input $\dot{\xi}^*$ as in Eq. \eqref{eq:input_momentum}, which gives 
   \begin{equation}
   \label{lyapunov-derivative-integral}
      \dot{V} = - \tilde{H}^\top K_d\tilde{H} - \zeta^\top K_o \zeta \leq 0. 
   \end{equation}
   This result implies stability of the equilibrium point and the boundedness of system's trajectories. Furthermore, as long as Eq. \eqref{eq:input_momentum_definition} holds, the closed loop dynamics is given by 
   \[\dot{\zeta} = - \zeta -K_o^{-1}\tilde{H},\] 
   and Eq. \eqref{eq:dot_I}-\eqref{eq:dot_tilde_H}. The system is therefore autonomous, and the convergence of $\tilde{H}, \zeta$ and $\dot{\tilde{H}}, \dot{\zeta}$  to zero can be proved via the La Salle's theorem. Convergence to zero of $I$ can be proven by computing Eq. \eqref{eq:dot_tilde_H} on the invariant manifold.
   
   Analogously to the proof of Lemma \ref{lemma_12},
   we are left to show that in a neighborhood of the equilibrium point, the variable $\xi$ remains bounded, which guarantees that \eqref{eq:input_momentum_definition} holds. Note that since the desired momentum $H_d$ is a feasible equilibrium, then there exists a bounded $\xi_e(t)$ such that 
  \[\dot{H}_d = A\phi(\xi_e) -mge_3.\]
   Consider now the right hand side of \eqref{eq:dot_tilde_H}, namely
\begin{IEEEeqnarray}{LCL}
Af -mge_3 -\dot{H}_d = \zeta - K_d \tilde{H} - K_p I,  \label{eq:dot_tilde_H_1}
\end{IEEEeqnarray}
   and linearize $Af -mge_3 -\dot{H}_d$  close to the equilibrium trajectory $\xi_e$, which yields
\begin{IEEEeqnarray}{LCL}
\label{linearisation2}
A\Phi(\xi_e)\tilde{\xi} &=& \zeta - K_d \tilde{H} - K_p I = \Bar{K} 
\begin{pmatrix}
I \\ \tilde{H} \\ \zeta
\end{pmatrix}
\label{eq:dot_tilde_H_2}
\end{IEEEeqnarray}
with $\Bar{K} = (-K_p, \ -K_d, \ 1_6)$. Since it is assumed that the robot makes a single contact with the environment, namely $n_c = 1$, then the matrix $A \in \mathbb{R}^{6\times6}$. From \eqref{matrix_Ak} one easily verifies that the matrix $A$ is always invertible, with $A^{-1}$ a bounded matrix. Then, one has 
\[|\tilde{\xi}| = \left|(A\Phi(\xi_e))^{-1}\Bar{K}
\begin{pmatrix}
I \\ \tilde{H} \\ \zeta
\end{pmatrix}\right| \leq c_b\left|
\begin{pmatrix}
I \\ \tilde{H} \\ \zeta
\end{pmatrix}\right|,\]
with $c_b > 0$. As a consequence of the Lyapunov stability of $(I,\tilde{H},\zeta) = (0,0,0)$,  there exist initial conditions $(I,\tilde{H},\zeta)(0)$ for which the variable $\tilde{\xi}$ stays bounded and as close as we wish to zero, thus rendering the linearisation \eqref{linearisation2} consistent $\forall \ t$. Finally, since $\xi_e$ is bounded, then $\xi$ is also bounded, and the equality \eqref{eq:input_momentum_definition} can be satisfied $\forall t$.
   
 
 \noindent
 \textbf{Proof of 2):} it is assumed that the  $\xi$ is bounded $\forall t$; then \eqref{eq:input_momentum_definition} is satisfied $\forall \ t$ because the matrix $A\Phi(\xi)$ in \eqref{eq:input_momentum_definition} remains of full rank. In view of the radial unboundedness of \eqref{lyapunov-integral}, and of the expression of \eqref{lyapunov-derivative-integral} that is valid $\forall t$, then global asymptotic stability follows from the same La Salle arguments above. 
   

   \subsection{Computation of $\dot{\xi}_0$:} 
   \label{computation-xi-0}
   The calculation of $\dot{\xi}_0$ is carried out by substituting \eqref{eq:input_momentum} 
   in the expression of \eqref{eq:input_torques}. Fist, rewrite \eqref{eq:input_torques} as:
   \begin{subequations} 
      \label{input_compact}
      \begin{align}
        \tau = \,\, &  \Theta f + \theta\\
        \Theta:= \,\, & -\Lambda^{\dagger}JM^{-1}J^\top\\
        \theta:= \,\, &  \Lambda^{\dagger}[JM^{-1}h - \dot{J}\nu] + N_{\Lambda}\tau_0.
      \end{align}
   \end{subequations}
   Then, consider the following Lyapunov function:
   \begin{IEEEeqnarray}{LCL}
     V  =  \frac{1}{2}\tau^{\top}\tau \nonumber \quad \Rightarrow \quad
     \dot{V}  =  \tau^{\top}\dot{\tau} = \tau^{\top}(\dot{\Theta}f + \Theta\Phi\dot{\xi} + \dot{\theta}). \nonumber
   \end{IEEEeqnarray}
   Substitute now $\dot{\xi}$ given by \eqref{eq:input_momentum} into 
   $\dot{V}$,
   which leads to:
   \begin{IEEEeqnarray}{LCL}
     \label{eq:lyap_xi_0}
     \dot{V} & = & \tau^{\top}(\dot{\Theta}f + \Theta\Phi\dot{\xi}_1 + \Theta\Phi N_{A\Phi}\dot{\xi}_0 + \dot{\theta}),
   \end{IEEEeqnarray}
    $\dot{\xi}_1 {:=} {(A\Phi)}^{\dagger}\,[\ddot{H}_d {-} (K_d {+} 1_6)\dot{\tilde{H}} {-} (K_d {+} K^{-1}_o {+} K_p)\tilde{H} {-} K_p I {-}\dot{A}f]$. A solution that minimizes the joint torque norm is to impose:
   \begin{IEEEeqnarray}{LCL}
     \label{eq:input_xi_0}
     \dot{\Theta}f + \Theta\Phi\dot{\xi}_1 + \Theta\Phi N_{A\Phi}\dot{\xi}_0 + \dot{\theta} = -K_{\tau}\tau, 
   \end{IEEEeqnarray}
   with $K_{\tau}$ a symmetric and positive definite matrix. When the equivalence \eqref{eq:input_xi_0} is satisfied, the Lyapunov derivative Eq. \eqref{eq:lyap_xi_0} becomes $\dot{V} = -\tau^{\top}K_{\tau}\tau \leq 0$ and the input joint torques converge to zero. However, this is not the case as the rank of the matrix $(\Theta\Phi N_{A\Phi})$ that multiplies the free variable $\dot{\xi}_0$ is lower than the dimension of the joint torques vector $\tau \in \mathbb{R}^n$. Nevertheless, we compute the closest solution to Eq. \eqref{eq:input_xi_0}, that leads to the following expression of $\dot{\xi}_0$:
   \begin{IEEEeqnarray}{LCL}
     \dot{\xi}_0 = -(\Theta\Phi N_{A\Phi})^{\dagger}(\dot{\Theta}f + \Theta\Phi\dot{\xi}_1 + \dot{\theta} +K_{\tau}\tau). \nonumber
   \end{IEEEeqnarray}

%% file: main.bbl
\begin{thebibliography}{10}
\providecommand{\url}[1]{#1}
\csname url@samestyle\endcsname
\providecommand{\newblock}{\relax}
\providecommand{\bibinfo}[2]{#2}
\providecommand{\BIBentrySTDinterwordspacing}{\spaceskip=0pt\relax}
\providecommand{\BIBentryALTinterwordstretchfactor}{4}
\providecommand{\BIBentryALTinterwordspacing}{\spaceskip=\fontdimen2\font plus
\BIBentryALTinterwordstretchfactor\fontdimen3\font minus
  \fontdimen4\font\relax}
\providecommand{\BIBforeignlanguage}[2]{{%
\expandafter\ifx\csname l@#1\endcsname\relax
\typeout{** WARNING: IEEEtran.bst: No hyphenation pattern has been}%
\typeout{** loaded for the language `#1'. Using the pattern for}%
\typeout{** the default language instead.}%
\else
\language=\csname l@#1\endcsname
\fi
#2}}
\providecommand{\BIBdecl}{\relax}
\BIBdecl

\bibitem{deWit1996}
\BIBentryALTinterwordspacing
C.~C. de~Wit, B.~Siciliano, and G.~Bastin, \emph{Motion and force
  control}.\hskip 1em plus 0.5em minus 0.4em\relax London: Springer London,
  1996, pp. 141--175. [Online]. Available:
  \url{https://doi.org/10.1007/978-1-4471-1501-4_4}
\BIBentrySTDinterwordspacing

\bibitem{yoshikawa2000}
T.~{Yoshikawa}, ``Force control of robot manipulators,'' in \emph{Proceedings
  2000 ICRA. Millennium Conference. IEEE International Conference on Robotics
  and Automation. Symposia Proceedings (Cat. No.00CH37065)}, vol.~1, April
  2000, pp. 220--226 vol.1.

\bibitem{mistry2010}
M.~{Mistry}, J.~{Buchli}, and S.~{Schaal}, ``Inverse dynamics control of
  floating base systems using orthogonal decomposition,'' in \emph{2010 IEEE
  International Conference on Robotics and Automation}, May 2010, pp.
  3406--3412.

\bibitem{sentis2005}
L.~{Sentis} and O.~{Khatib}, ``Control of free-floating humanoid robots through
  task prioritization,'' in \emph{Proceedings of the 2005 IEEE International
  Conference on Robotics and Automation}, April 2005, pp. 1718--1723.

\bibitem{Frontiers2015}
F.~Nori, S.~Traversaro, J.~Eljaik, F.~Romano, A.~Del~Prete, and D.~Pucci,
  ``i{C}ub whole-body control through force regulation on rigid noncoplanar
  contacts,'' \emph{Frontiers in Robotics and AI}, vol.~2, no.~6, pp. 1--18,
  2015.

\bibitem{Villani2015}
\BIBentryALTinterwordspacing
L.~Villani, \emph{Force Control in Robotics}.\hskip 1em plus 0.5em minus
  0.4em\relax London: Springer London, 2015, pp. 463--469. [Online]. Available:
  \url{https://doi.org/10.1007/978-1-4471-5058-9_169}
\BIBentrySTDinterwordspacing

\bibitem{ortenzi2017}
\BIBentryALTinterwordspacing
V.~Ortenzi, R.~Stolkin, J.~Kuo, and M.~Mistry, ``Hybrid motion/force control: a
  review,'' \emph{Advanced Robotics}, vol.~31, no. 19-20, pp. 1102--1113, 2017.
  [Online]. Available: \url{https://doi.org/10.1080/01691864.2017.1364168}
\BIBentrySTDinterwordspacing

\bibitem{raibert1981}
M.~H. Raibert and J.~J. Craig, ``Hybrid position/force control of
  manipulators,'' in \emph{Journal of Dynamic Systems, Measurement, and
  Control}, June 1981.

\bibitem{Featherstone2007}
R.~Featherstone, \emph{Rigid Body Dynamics Algorithms}.\hskip 1em plus 0.5em
  minus 0.4em\relax Secaucus, NJ, USA: Springer-Verlag New York, Inc., 2007.

\bibitem{Ott2011}
C.~Ott, M.~Roa, and G.~Hirzinger, ``Posture and balance control for biped
  robots based on contact force optimization,'' in \emph{Humanoid Robots
  (Humanoids), 2011 11th IEEE-RAS International Conference on}, Oct 2011, pp.
  26--33.

\bibitem{Wensing2013}
P.~Wensing and D.~Orin, ``Generation of dynamic humanoid behaviors through
  task-space control with conic optimization,'' in \emph{Robotics and
  Automation (ICRA), 2013 IEEE International Conference on}, May 2013, pp.
  3103--3109.

\bibitem{Hopkins2015a}
M.~Hopkins, R.~Griffin, A.~Leonessa, B.~Lattimer, and T.~Furukawa, ``Design of
  a compliant bipedal walking controller for the darpa robotics challenge,'' in
  \emph{Humanoid Robots (Humanoids), 2015 IEEE-RAS 15th International
  Conference on}, Nov 2015, pp. 831--837.

\bibitem{spong1998}
M.~W. Spong, ``Underactuated mechanical systems,'' in \emph{Control Problems in
  Robotics and Automation}, B.~Siciliano and K.~P. Valavanis, Eds.\hskip 1em
  plus 0.5em minus 0.4em\relax Berlin, Heidelberg: Springer Berlin Heidelberg,
  1998, pp. 135--150.

\bibitem{Azad2015}
M.~{Azad} and M.~N. {Mistry}, ``Balance control strategy for legged robots with
  compliant contacts,'' in \emph{2015 IEEE International Conference on Robotics
  and Automation (ICRA)}, May 2015, pp. 4391--4396.

\bibitem{Caron2019}
S.~{Caron}, A.~{Escande}, L.~{Lanari}, and B.~{Mallein}, ``Capturability-based
  pattern generation for walking with variable height,'' \emph{IEEE
  Transactions on Robotics}, pp. 1--20, 2019.

\bibitem{Henze2018}
B.~{Henze}, R.~{Balachandran}, M.~A. {Roa-Garzón}, C.~{Ott}, and
  A.~{Albu-Schäffer}, ``Passivity analysis and control of humanoid robots on
  movable ground,'' \emph{IEEE Robotics and Automation Letters}, vol.~3, no.~4,
  pp. 3457--3464, Oct 2018.

\bibitem{liu2015b}
M.~{Liu} and V.~{Padois}, ``Reactive whole-body control for humanoid balancing
  on non-rigid unilateral contacts,'' in \emph{2015 IEEE/RSJ International
  Conference on Intelligent Robots and Systems (IROS)}, Sep. 2015, pp.
  3981--3987.

\bibitem{park2013}
H.~{Park}, A.~{Ramezani}, and J.~W. {Grizzle}, ``A finite-state machine for
  accommodating unexpected large ground-height variations in bipedal robot
  walking,'' \emph{IEEE Transactions on Robotics}, vol.~29, no.~2, pp.
  331--345, April 2013.

\bibitem{liu2015}
M.~Liu, R.~Lober, and V.~Padois, ``Whole-body hierarchical motion and force
  control for humanoid robots,'' \emph{Autonomous Robots}, vol.~40, 10 2015.

\bibitem{mansard2009}
N.~{Mansard}, O.~{Stasse}, P.~{Evrard}, and A.~{Kheddar}, ``A versatile
  generalized inverted kinematics implementation for collaborative working
  humanoid robots: The stack of tasks,'' in \emph{2009 International Conference
  on Advanced Robotics}, June 2009, pp. 1--6.

\bibitem{koolen2015design}
T.~Koolen, S.~Bertrand, G.~Thomas, T.~de~Boer, T.~Wu, J.~Smith, J.~Englsberger,
  and J.~Pratt, ``Design of a momentum-based control framework and application
  to the humanoid robot atlas,'' \emph{International Journal of Humanoid
  Robotics}, vol.~13, p.~34, March 2016.

\bibitem{nava2016}
G.~Nava, F.~Romano, F.~Nori, and D.~Pucci, ``Stability analysis and design of
  momentum-based controllers for humanoid robots,'' in \emph{2016 IEEE/RSJ
  International Conference on Intelligent Robots and Systems (IROS)}, Oct 2016,
  pp. 680--687.

\bibitem{Lee2012}
\BIBentryALTinterwordspacing
S.-H. Lee and A.~Goswami, ``A momentum-based balance controller for humanoid
  robots on non-level and non-stationary ground,'' \emph{Autonomous Robots},
  vol.~33, no.~4, pp. 399--414, Nov 2012. [Online]. Available:
  \url{https://doi.org/10.1007/s10514-012-9294-z}
\BIBentrySTDinterwordspacing

\bibitem{traversaro2017}
\BIBentryALTinterwordspacing
S.~Traversaro, D.~Pucci, and F.~Nori, ``A unified view of the equations of
  motion used for control design of humanoid robots: the role of the base frame
  in free-floating mechanical systems and its connection to centroidal
  dynamics,'' 2017. [Online]. Available:
  \url{https://www.semanticscholar.org/paper/A-Unified-View-of-the-Equations-of-Motion-used-for-Traversaro-Pucci/fa674bbe84cb6b0b85c4a351582bd2a66fd8df59}
\BIBentrySTDinterwordspacing

\bibitem{Marsden1999}
J.~E. Marsden and T.~S. Ratiu, \emph{Lagrangian Mechanics}.\hskip 1em plus
  0.5em minus 0.4em\relax New York, NY: Springer New York, 1999, pp. 181--218.

\bibitem{dafarra2018}
S.~{Dafarra}, G.~{Nava}, M.~{Charbonneau}, N.~{Guedelha}, F.~{Andrade},
  S.~{Traversaro}, L.~{Fiorio}, F.~{Romano}, F.~{Nori}, G.~{Metta}, and
  D.~{Pucci}, ``A control architecture with online predictive planning for
  position and torque controlled walking of humanoid robots,'' in \emph{2018
  IEEE/RSJ International Conference on Intelligent Robots and Systems (IROS)},
  Oct 2018, pp. 1--9.

\bibitem{bouyarmane2018}
K.~{Bouyarmane}, K.~{Chappellet}, J.~{Vaillant}, and A.~{Kheddar}, ``Quadratic
  programming for multirobot and task-space force control,'' \emph{IEEE
  Transactions on Robotics}, vol.~35, no.~1, pp. 64--77, Feb 2018.

\bibitem{marie2016}
M.~{Charbonneau}, F.~{Nori}, and D.~{Pucci}, ``On-line joint limit avoidance
  for torque controlled robots by joint space parametrization,'' in \emph{2016
  IEEE-RAS 16th International Conference on Humanoid Robots (Humanoids)}, Nov
  2016, pp. 899--904.

\bibitem{isidori2013}
\BIBentryALTinterwordspacing
A.~Isidori, ``The zero dynamics of a nonlinear system: From the origin to the
  latest progresses of a long successful story,'' \emph{European Journal of
  Control}, vol.~19, no.~5, pp. 369 -- 378, 2013, the Path of Control.
  [Online]. Available:
  \url{http://www.sciencedirect.com/science/article/pii/S0947358013000836}
\BIBentrySTDinterwordspacing

\bibitem{carpentier2018analytical}
J.~Carpentier and N.~Mansard, ``Analytical derivatives of rigid body dynamics
  algorithms,'' in \emph{Robotics: Science and Systems}, 2018, pp. 1--10.

\bibitem{Andersson2018}
J.~A.~E. Andersson, J.~Gillis, G.~Horn, J.~B. Rawlings, and M.~Diehl,
  ``{CasADi} -- {A} software framework for nonlinear optimization and optimal
  control,'' \emph{Mathematical Programming Computation}, vol.~11, pp. 1--36,
  March 2019.

\bibitem{orin08}
D.~E. Orin and A.~Goswami, ``Centroidal momentum matrix of a humanoid robot:
  Structure and properties,'' \emph{Intelligent Robots and Systems, 2008. IROS
  2008. IEEE/RSJ International Conference on}, pp. 653 -- 659, 2008.

\bibitem{Metta20101125}
G.~Metta, L.~Natale, F.~Nori, G.~Sandini, D.~Vernon, L.~Fadiga, C.~von Hofsten,
  K.~Rosander, M.~Lopes, J.~Santos-Victor, A.~Bernardino, and L.~Montesano,
  ``The i{C}ub humanoid robot: An open-systems platform for research in
  cognitive development,'' \emph{Neural Networks}, vol.~23, no. 8–9, pp. 1125
  -- 1134, 2010, social Cognition: From Babies to Robots.

\bibitem{Koenig04}
N.~Koenig and A.~Howard, ``Design and use paradigms for gazebo, an open-source
  multi-robot simulator,'' \emph{Intelligent Robots and Systems, 2004. (IROS
  2004). Proceedings. 2004 IEEE/RSJ International Conference on}, pp. 2149 --
  2154, 2004.

\end{thebibliography}
